%% file: main.tex
\PassOptionsToPackage{table, dvipsnames}{xcolor}
\documentclass[conference,compsoc]{IEEEtran}

\ifCLASSOPTIONcompsoc
  \usepackage[nocompress]{cite}
\else
  \usepackage{cite}
\fi

\ifCLASSINFOpdf
\else
\fi

\input{mydefs}

\begin{document}

\date{}

\title{\Large \bf Dialogue Injection Attack: Jailbreaking LLMs through Context Manipulation}

\author{
Wenlong Meng\IEEEauthorrefmark{1},
Fan Zhang\IEEEauthorrefmark{1},
Wendao Yao\IEEEauthorrefmark{1},
Zhenyuan Guo\IEEEauthorrefmark{1},
Yuwei Li\IEEEauthorrefmark{2},
Chengkun Wei\IEEEauthorrefmark{1}\IEEEauthorrefmark{4},
Wenzhi Chen\IEEEauthorrefmark{1}
\\
\IEEEauthorrefmark{1}Zhejiang University, \IEEEauthorrefmark{2}National University of Defense Technology
\\
\{mengwl, zhenyuanguo, weichengkun, chenwz\}@zju.edu.cn,\\ \{fan1.22, wendao.22\}@intl.zju.edu.cn, liyuwei@nudt.edu.cn
}

\maketitle
\begin{NoHyper}
\def\thefootnote{\IEEEauthorrefmark{4}}\footnotetext{Corresponding author.}
\end{NoHyper}

\begin{abstract}
Large language models (LLMs) have demonstrated significant utility in a wide range of applications; however, their deployment is plagued by security vulnerabilities, notably jailbreak attacks.
These attacks manipulate LLMs to generate harmful or unethical content by crafting adversarial prompts.
While much of the current research on jailbreak attacks has focused on single-turn interactions, it has largely overlooked the impact of historical dialogues on model behavior.
In this paper, we introduce a novel jailbreak paradigm, Dialogue Injection Attack (\method), which leverages the dialogue history to enhance the success rates of such attacks.
\method operates in a black-box setting, requiring only access to the chat API or knowledge of the LLM’s chat template.
We propose two methods for constructing adversarial historical dialogues: one adapts gray-box prefilling attacks, and the other exploits deferred responses.
Our experiments show that \method achieves state-of-the-art attack success rates on recent LLMs, including Llama-3.1 and GPT-4o. Additionally, we demonstrate that DIA can bypass 5 different defense mechanisms, highlighting its robustness and effectiveness.\footnote{Code is available at \url{https://github.com/meng-wenlong/DIA}.}
\end{abstract}

\mypara{Disclaimer} 
This paper contains model outputs that may be considered offensive to readers.

\section{Introduction}

The advent of \textit{large language models} (LLMs) such as GPT-4~\cite{achiam2023gpt} and PaLM~\cite{chowdhery2023palm} have revolutionized the field of natural language processing, exhibiting great power in natural language generation.
Due to their robust generative abilities, LLMs have been deployed across a multitude of content generation applications, including code generators~\cite{gu2023llm}, translators~\cite{koshkin2024transllama}, and personalized recommendation systems~\cite{lin2024data}.
Among these, chatbots represent the most popular application, exemplified by OpenAI ChatGPT and Google Bard.
Since LLMs have been trained on a broad corpus of Web data, which includes both compliant and non-compliant content, they might generate harmful content when prompted~\cite{dai2024bias, xu2024pride}.
To mitigate these risks, LLM developers employ alignment techniques such as \textit{reinforcement learning from human feedback} (RLHF)~\cite{ouyang2022training} to ensure outputs conform to human ethical standards.
This alignment guarantees that LLMs like ChatGPT adhere to ethical norms and foster constructive engagement.

However, the security mechanisms integrated into LLMs are not foolproof.
The threat of \textit{jailbreak attacks} persists, where adversaries craft specialized prompts to coax LLMs into producing harmful or unethical outputs~\cite{shen2023anything, yu2024don}.
These attacks employ various techniques such as hypnosis~\cite{li2023deepinception}, sensitive words disguise~\cite{liu2024making}, and tense shift~\cite{andriushchenko2024does}.
Such manipulated outputs could facilitate further malicious activities, including the extraction of personal information from training data~\cite{li2023multi} or the creation of malware~\cite{lin2024malla}.
Studying these jailbreak attacks is crucial for developers to enhance the security of LLM applications and improve their alignment with safety standards.

Current research on jailbreak methodologies predominantly focuses on single-turn interactions, where all adversarial content is compressed into a single prompt.
However, the influence of dialogue history plays a significant role in shaping the outputs of LLMs, especially in chatbot applications.
Existing LLM chatbots utilize a \textit{chat template} to concatenate historical dialogues with the current prompt before submitting it to the backend LLM for continuation.
These historical dialogues act as part of the input.
The key challenge of exploiting historical dialogues is that historical assistant text is usually considered unmanipulatable and cannot be altered by the attacker.
Given the significant impact of historical dialogues on an LLM’s responses, we explore the research question:
\textit{How can historical dialogue manipulation be made practical, and how can it be utilized to enhance jailbreak attacks?}

\mypara{Our Contributions}
Our research takes the first step in exploring how historical dialogues can be used to enhance jailbreaking attacks.
We introduce a novel black-box jailbreak paradigm named \method (\underline{D}ialogue \underline{I}njection \underline{A}ttack), which leverages the structural design of LLMs' chat templates to inject deceptive dialogues into the input.
We demonstrate that an attacker, knowing the LLM's chat template, can use the structure of the template to insert arbitrary historical dialogues into the input (\autoref{sec:response_manipulation}), including assistant text and system text, which is usually considered unrealistic in black-box scenarios.
Furthermore, we show how attackers can reduce the requirement of chat template knowledge by employing a \textit{template inference attack} to deduce which chat template the victim LLM is using (\autoref{sec:template_inference_attack}).
Our \method provides a novel attack vector to jailbreak by constructing a deceptive dialogue history.

Based on the \method paradigm, we propose two methods for constructing adversarial dialogues, namely \method-I and \method-II.
\method-I is based on \textit{prefilling attacks}, which place an affirmative prefix at the beginning of the assistant's text.
We adapt this approach to black-box scenarios using a ``continue'' command and design an automated module to generate affirmative beginnings.
\method-II is based on a new finding that deferred malicious responses have a higher log-likelihood than those immediately following its prompt.
We persuade the LLM to perform a word substitution task, achieving both answer deferral and malicious content disguise.
In addition, we develop a prompt rewrite algorithm to enhance prompt diversity in multi-query scenarios.
After 10 queries, our \method achieves an attack success rate of 0.89 on Llama-3.1-8B and 0.82 on GPT-4o evaluated on AdvBench.

In summary, our contributions are outlined as follows:
\begin{icompact}

\item
We present a novel black-box jailbreak attack paradigm that leverages historical dialogues to enhance jailbreak attack performance.
We provide the historical dialogue manipulating method that can inject arbitrary attacker-crafted dialogues, with the only requirement of victim LLMs' chat template.
Furthermore, we introduce the template inference attack to infer LLMs' chat templates.

\smallskip
\item 
Building on our proposed paradigm, we develop two methods for constructing adversarial dialogues: \method-I and \method-II.
\method-I adapts prefilling attacks to black-box scenarios, while \method-II employs word substitution tasks to delay the delivery of actual answers.
Additionally, we introduce a prompt rewrite algorithm designed to enhance the effectiveness of multi-query attacks.

\smallskip
\item
We evaluate \method-I and \method-II across three open-source jailbreak benchmarks against 10 LLMs, comparing their effectiveness with four state-of-the-art black-box jailbreak attacks.
The experimental results demonstrate that our attacks achieved superior attack success rates on recent LLMs, e.g., 0.89 on Llama-3.1-8B and 0.82 on GPT-4o as tested on AdvBench.
We also assess the robustness of \method-I and \method-II in penetrating defenses.
Among the five defenses evaluated, \method-I achieves an average defense pass rate of 0.93, while \method-II achieves 0.82.

\end{icompact}

\section{Preliminaries}

\subsection{Large Language Models}
Morden LLMs, like GPT-4~\cite{achiam2023gpt}, Llama~\cite{touvron2023llama}, Gemma~\cite{team2024gemma} etc., are decoder-only Transformer models trained on next token prediction tasks.
These models consist of a stack of layers identical in structure.
Each layer contains two modules: self-attention and feed-forward network (FFN).
The training of LLMs consists of two phases: \textit{pre-training} and \textit{post-training}.

In the pre-training phase, the model is trained on a vast corpus of text data, learning to understand and generate language by predicting the next words. The pre-training phase is unsupervised, meaning it does not require labeled data, and its dataset is usually crawled from the Internet.
Given the sequence $X = \{ x_t \}_{1}^{T}$, the pre-training loss for an LLM $\pi\left( \cdot ; \theta \right)$ is
$$
\mathcal{L}=-\sum_{t=1}^T \log \pi\left(x_t \mid x_{<t} ; \theta\right).
$$
The post-training phase is designed to enhance instruction-following capabilities.
In this phase, the training data is divided into prompts and responses, which are alternately concatenated by a chat template into a continuous text block.
Typically, the post-training loss is only computed on the final response~\cite{rafailov2024direct, hong2024orpo}.

\begin{figure}[!tbp]
\centering
\includegraphics[width=\columnwidth]{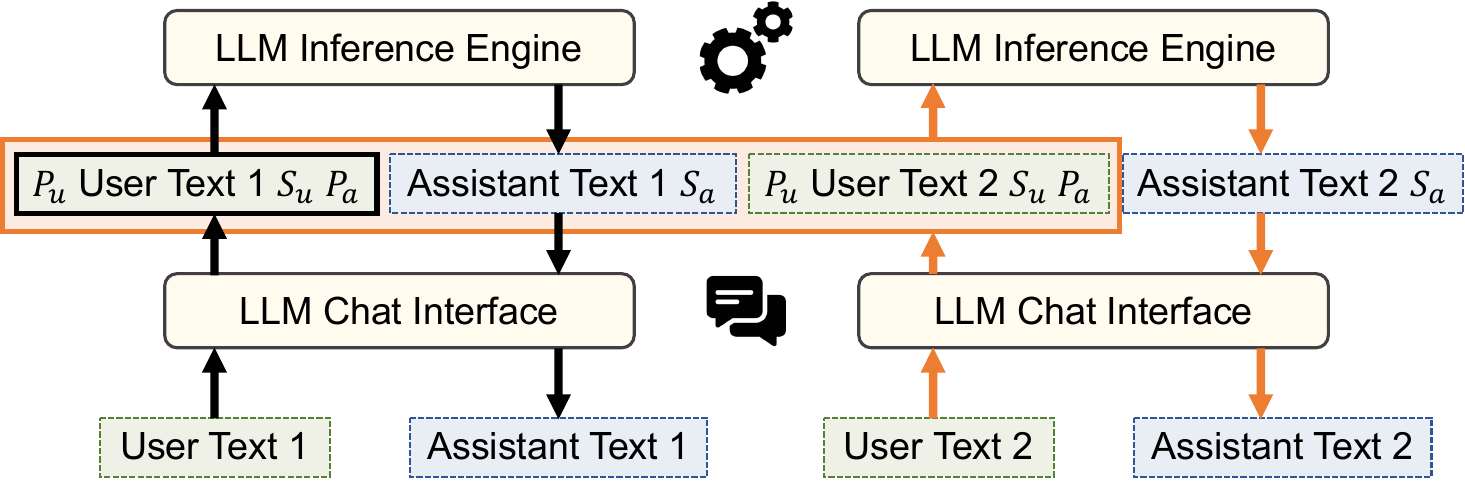}
\caption{LLM chat pipeline.}
\label{fig:chat_pipeline}
\vspace{-0.3cm}
\end{figure}

\begin{table*}[!htbp]
\centering
\caption{Examples of LLM chat templates structure. Llama-3 and Gemma-2 use different prefixes to indicate the roles.}
\label{tab:chat_template_structure}
\begin{tabular}{clll}
    \toprule
    \textbf{Component} & \textbf{Description} & \textbf{Llama-3} & \textbf{Gemma-2} \\
    \midrule
    $P_s$ & Prefix of system text & \texttt{<|start\_header\_id|>system<|end\_header\_id|>\textbackslash n\textbackslash n} & - \\
    $S_s$ & Suffix of system text & \texttt{<|eot\_id|>} & - \\
    $P_u$ & Prefix of user text & \texttt{<|start\_header\_id|>user<|end\_header\_id|>\textbackslash n\textbackslash n} & \texttt{<start\_of\_turn>user\textbackslash n} \\
    $S_u$ & Suffix of user text & \texttt{<|eot\_id|>} & \texttt{<end\_of\_turn>\textbackslash n} \\
    $P_a$ & Prefix of assistant text & \texttt{<|start\_header\_id|>assistant<|end\_header\_id|>\textbackslash n\textbackslash n} & \texttt{<start\_of\_turn>model\textbackslash n} \\
    $S_a$ & Suffix of assistant text & \texttt{<|eot\_id|>} & \texttt{<end\_of\_turn>\textbackslash n} \\
    \bottomrule
\end{tabular}
\vspace{-0.3cm}
\end{table*}

\subsection{LLM Chat Pipeline}
\label{sec:llm_chat_pipeline}

The fundamental ability of LLMs is to predict the next word according to previous input words.
To handle chat tasks, LLMs utilize a chat template to integrate historical dialogues with the user prompt into a continuous text.
Different types of LLMs often employ distinct chat templates.
Despite their variations, modern chat templates typically adhere to a similar structure, wherein prefixes and suffixes are added to the text corresponding to different roles.
For instance, Llama-3 and Gemma-2 exemplify this practice, as shown in~\autoref{tab:chat_template_structure}.
In chat tasks, the roles generally include a user and an assistant, with many models (e.g., Llama-3 and Llama-2) also introducing a system entity to prioritize settings for the LLM.
These chat templates assign distinct prefixes or suffixes to the text from different roles to indicate the originator of each message.
Using a chat template, a set of dialogue messages can be converted to the following text:
\begin{equation}
\begin{split}
X = &P_s\langle\text {system\_text}\rangle S_s \\
&\left[P_u\langle\text {user\_text}\rangle S_u P_a\langle\text {assistant\_text}\rangle S_a\right] \times n,\nonumber
\end{split}
\end{equation}
where $n$ represents $n$ rounds of conversation.

An LLM chat service comprises a chat interface as the frontend and an inference engine as the backend. The chat interface receives user prompts, wraps them with a chat template, and then sends them to the inference engine.
After the inference engine generates the response, the chat interface then delivers the results back to the user.
\autoref{fig:chat_pipeline} depicts how a chat service processes users' requests.
A chat interface could be a WebUI (e.g., OpenAI ChatGPT or Google Gemini) or an API (e.g., OpenAI API or Google Gemini API).
An inference engine, like vLLM~\cite{kwon2023efficient} or Ollama~\cite{ollama}, is designed to meet latency-related Service Level Objectives (SLOs).
Modern LLM inference engines boost inference efficiency by separating inference into prefill and decode stages and employing the Key-Value (KV) cache to minimize redundant computations~\cite{lee2024infinigen, qin2024mooncake}.
In~\autoref{fig:chat_pipeline}, during the inference stage of the second round chat, the prompt and response from the first round are concatenated to user text in the second round, ensuring the LLM has the memory of the previous conversion.

\subsection{Safty Alignment and Jailbreak Attacks}

\mypara{Safty Alignment} Due to exposure to a broad range of texts during pre-training, including both safe and unsafe sources, LLMs can generate unethical content.
To ensure the safety of the content produced by LLMs, developers often incorporate safety alignment during the post-training phase.
Common safety alignment methods include Supervised Fine-Tuning (SFT), Reinforcement Learning from Human Feedback (RLHF)~\cite{ouyang2022training}, and Direct Preference Optimization (DPO)~\cite{rafailov2024direct}.
The goal of these methods is to enable LLMs to refuse to respond to harmful inquiries while still effectively answering benign questions.

\mypara{Jailbreak Attacks}
Jailbreak attacks refer to techniques that exploit vulnerabilities in LLMs to bypass their built-in safety and ethical guidelines.
These attacks aim to manipulate the model's behavior, enabling it to generate harmful, biased, or otherwise restricted content that it is designed to avoid.
Depending on the threat model, existing jailbreak attacks can be classified into: \textit{white-box}, \textit{gray-box}, and \textit{black-box} categories. White-box attacks require access to the architecture and parameters of the victim LLM for training purposes.
For example, GCG~\cite{zou2023universal} optimizes an adversarial suffix through a combination of gradient heuristics and greedy search.
Gray-box attacks also need access to the model but do not require training.
A common type of gray-box attack is the prefilling attack~\cite{andriushchenko2024jailbreaking, jiang2024chatbug, lv2024adappa}, which simply prefills an LLM to start its response with an affirmative beginning, such as ``\textit{Sure, I'd be happy to help...}''.
Since general LLM chat services do not allow users to manipulate model responses directly, prefilling attacks require the attacker to perform inference locally.
Black-box attacks, on the other hand, do not require any knowledge of the victim model.
Current black-box jailbreak attacks typically use carefully crafted prompts to exploit vulnerabilities in safety alignment.
As developers cannot address all potential harmful scenarios during safety alignment, there remain backdoors that can circumvent safety guidelines and trigger harmful behaviors.
For instance, DeepInception~\cite{li2023deepinception} composes virtual scenes to hypnotize LLMs, while DRA~\cite{liu2024making} conceals harmful instructions through disguise and prompts the LLM to reconstruct the original harmful instruction within its completion.

\mypara{Limitations of Existing Jailbreak Attacks}
White-box and gray-box attacks require access to a model's parameters, which is often impractical, especially for proprietary models.
Additionally, these types of attacks demand significant computational power from attackers, limiting their widespread use and harm.
Although black-box jailbreak attacks are easier to perform, they are easier to be patched by RLHF.
After the expose of black-box jailbreak attacks, LLM developer can add these jailbreak prompts to their alignment dataset, which can help the LLM detect and thwart such attacks.
In~\autoref{sec:single_turn_attack}, we observe that two state-of-the-art black-box jailbreak attacks, DeepInception and DRA, experience ASR degradation by 67\% and 99\% on Llama-3.1-8B, respectively, compared to their performance on Llama-2-7B.

\section{Attack Intuition}
\label{sec:motivation}

\mypara{Motivation}
We observe that a major limitation of existing black-box jailbreak attacks is their focus on single-round conversions, where the goal is to jailbreak the LLM using only one round of dialogue.
However, as discussed in~\autoref{sec:llm_chat_pipeline}, the current user prompt is concatenated with previous dialogues to form the complete input for inference. As a result, historical dialogue plays a crucial role in shaping the model’s response.
Inspired by this observation, we extend the scope to multi-turn jailbreak attacks, exploring how manipulating historical dialogues can enhance attack success.
Specifically, we argue that \textit{an LLM, when presented with the same user prompt but different historical dialogues, will exhibit variations in its output distribution}.
This hypothesis can be formalized as:
\begin{equation}
    \pi\left(X\mid H ; \theta\right) \neq \pi\left(X\mid H^{\prime} ; \theta\right),\nonumber
\end{equation}
where $X$ is the user prompt of current round, $H$ and $H^{\prime}$ are history dialogues, $H\neq H^{\prime}$.
Existing black-box jailbreak attacks typically assume $H$ is empty, neglecting the impact of history dialogues.

In this section, we explore the practicality of using historical dialogues to implement jailbreak attacks.
Our threat model is detailed in~\autoref{sec:threat_model}.
We then introduce the technique of dialogue injection in~\autoref{sec:response_manipulation}, which enables an attacker to inject arbitrary historical dialogues into the input with the requirement of the knowledge of the LLM's chat template.
Following this, we discuss how to relax this requirement through template inference attacks in~\autoref{sec:template_inference_attack}.

\subsection{Threat Model}
\label{sec:threat_model}

\mypara{Attacker's Goal}
We adhere to the traditional objective of jailbreak attacks~\cite{yu2024don}, which is to manipulate a victim LLM into responding harmfully and accurately to a malicious prompt.
This means that the LLM should not only avoid refusing to respond but also must provide a reasonable output that fulfills the requirements of the prompt.

\mypara{Attacker's Capability}
In this paper, we consider the strict black-box scenario where the adversary lacks access to the architecture, parameters, and training data of the victim model, rendering it impossible to reproduce the model.
The adversary can interact with the victim LLM using two methods: 1) via a chat API, or 2) through a WebUI. 
In the second case, we assume the adversary has knowledge of the victim LLM's chat template.
Nevertheless, this assumption can be relaxed if the victim LLM uses a publicly available chat template.
The adversary could use the template inference attack, as detailed in~\autoref{sec:template_inference_attack}, to deduce the specific template used.

\subsection{Historical Dialogue Manipulation}
\label{sec:response_manipulation}

\mypara{Dialogue Injection}
Manipulating historical dialogue is straightforward when adversaries have access to the chat API, as they can directly set previous assistant responses.
Typically, completely controlling responses through a WebUI is deemed impossible since LLMs are probabilistic models and their outputs cannot be predetermined.
However, we propose that an adversary could manipulate previous assistant responses completely using a technique we refer to as \textit{dialogue injection}.
This method is inspired by \textit{SQL injection}, where malicious SQL statements are inserted into an entry field to be executed.
By utilizing a WebUI as the entry point, an adversary can strategically inject crafted responses by surrounding them with $S_u P_a$ and $S_a P_u$.
Formally, if the adversary wants to insert a crafted conversation $\left[ \langle\text {user\_text}\rangle^1,  \langle\text {assistant\_text}\rangle^1 \right]$ before $\langle\text {user\_text}\rangle^2$, they can compose their user prompt as:
\begin{equation}
\begin{split}
    X_{adv} = &\langle\text {user\_text}\rangle^1 S_u P_a \langle\text {assistant\_text}\rangle^1 S_a P_u \\ &\langle\text {user\_text}\rangle^2.
    \label{eq:dialogue_injection}
\end{split}
\end{equation}
In this way, the LLM inference engine receives a context the same as the round 2 generation illustrated in~\autoref{fig:chat_pipeline}.
The difference is that while the assistant text in~\autoref{fig:chat_pipeline} is generated by the LLM, $\langle\text {assistant\_text}\rangle^1$ in~\autoref{eq:dialogue_injection} is fabricated by the attacker.
The attacker can also insert multiple conversations into the inputs using the methods outlined in~\autoref{eq:dialogue_injection}.
Thanks to advancements in expanding the input sequence length of LLMs~\cite{peng2023yarn, xiong2023effective, liu2023ring, tang2024razorattention}, current LLM chat WebUIs accept very long inputs, ranging from 2K tokens in ChatGPT to 100K in Claude~\cite{long-context}.
Thus, the attacker does not need to worry about exceeding input length limitations due to the insertion of dialogues.

By leveraging chat APIs or dialogue injection, certain gray-box jailbreak attacks can also be executed in black-box scenarios.
Below, we present three examples of such attacks.
Specifically, we demonstrate how these attacks can be achieved in black-box settings through the use of dialogue injection.

\mypara{Empowering Prefilling Attack}
This attack prefills the first few tokens with a non-refusal prefix at the beginning of the inference.
Prefilling attack exploits the characteristics of shallow safety alignment~\cite{qi2024safety}.
Safety alignment techniques often constrain only the probability distribution of the initial few tokens in an output.
Attackers can circumvent these alignments by setting the initial tokens of a response.
In black-box scenarios, we cannot alter the outputs of the LLM directly.
However, we can manipulate the model into believing it has already generated some content by injecting history dialogues. Then, we can make the LLM continue its previous output.
The corresponding adversarial prompt for the WebUI interface is
\begin{equation}
X_{adv} = \langle\text {user\_text}\rangle S_u P_a \langle\text {prefill}\rangle S_a P_u \langle \text {continue\_cmd} \rangle.\nonumber
\end{equation}

\mypara{Empowering System Prompt Replacement Attack}
System prompts have the highest priority.
If attackers can make the victim LLM lower its guard via a system prompt, they can increase the jailbreak success rate.
With access to the chat API, attackers can set their own system prompts.
As for WebUI access, attackers can also replace $\langle \text{system\_text} \rangle$ with $\langle \text{system\_text} \rangle^\prime$ using the adversarial input below
\begin{equation}
\begin{split}
X_{adv} = &\langle\text {user\_text}\rangle^1 S_u P_a \langle\text {assistant\_text}\rangle^1 S_a P_s \\
&\langle \text {system\_text} \rangle^\prime S_s P_a \langle\text {user\_text}\rangle^2.\nonumber
\end{split}
\end{equation}
Attackers can prompt the LLM to forget previous commands in $\langle \text{system\_text} \rangle^\prime$ to negate the effects of $\langle \text{system\_text} \rangle$.

\mypara{Empowering In-Context Demonstration Attack}
In-context learning has been shown to enhance the command-following capabilities of LLMs~\cite{dong2022survey, min2022rethinking}.
Recent studies also found that in-context demonstrations can increase jailbreak attack performance~\cite{cheng2024leveraging, wei2023jailbreak}.
Existing in-context jailbreak attacks require either gray box or chat API access.
Through dialogue injection, attackers can introduce arbitrary context via a WebUI
\begin{equation}
\begin{split}
X_{adv} = &[\langle\text {user\_text}\rangle^n S_u P_a \langle\text {assistant\_text}\rangle^n S_a P_u]_1^N \\
&\langle\text {user\_text}\rangle^{N+1}.\nonumber
\end{split}
\end{equation}

\subsection{Template Inference Attack}
\label{sec:template_inference_attack}
A practical concern with dialogue injection is that it requires the attacker to have knowledge of the chat template.
A chat template of an LLM can be arbitrary.
However, LLM developers generally use post-trained models for secondary development since post-training requires substantial datasets and computational resources.
It is challenging for individual developers and small companies to conduct high-quality post-training.
It is also difficult for LLM developers to change the chat template of a post-trained model because template transfer can lead to performance degradation.
In our experiments with the Gemma-2-2B model~\cite{team2024gemma}, we attempted a template transfer using 10,000 entries from the NoRobots dataset~\cite{no_robots}.
We input the user prompts from NoRobots into Gemma-2-2B, collected the outputs, and then concatenated these prompts and outputs using the Llama-3 template to fine-tune Gemma-2-2B for one epoch.
We observed that the AlpacaEval 2.0 score~\cite{dubois2024length} of Gemma-2-2B decreased from 41.2 to 28.8, indicating that LLM developers cannot trivially alter the chat templates of post-trained models.

If an LLM developer uses an open-sourced chat template, we propose a method to infer which chat template is used by the LLM, named \textit{template inference attack}.
Assume an LLM utilizes one chat template, $T_{inf}$, from an open-source collection of chat templates $\mathcal{T}=\{T_n\}_1^N$.
If we could determine whether $T_n=T_{inf}$, it would enable us to iterate through $\mathcal{T}$ to identify $T_{inf}$.
We propose a method to ascertain whether $T_i = T_{inf}$ or $T_j = T_{inf}$ by combining two templates, $T_i$ and $T_j$, to create a probe.
The probe prompt formed as:
\begin{equation}
\begin{split}
    X_{P} = &\langle\text {user\_text}\rangle S_{ui}P_{ai}S_{uj}P_{aj} \langle\text {assistant\_text}\rangle \\
    &S_{ai}P_{ui}S_{aj}P_{uj}\langle\text {repeat\_cmd}\rangle,
    \label{eq:probe}
\end{split}
\end{equation}
where $\langle\text {repeat\_cmd}\rangle$ is a command to ask the LLM to repeat its previous response.
When $T_i = T_{inf}$, $S_{ui}P_{ai}$ and $S_{ai}P_{ui}$ will act as the template part.
LLM will output $S_{uj}P_{aj} \langle\text {assistant\_text}\rangle$.
When $T_j = T_{inf}$, $S_{uj}P_{aj}$ and $S_{aj}P_{uj}$ will act as the template part.
LLM will output $\langle\text {assistant\_text}\rangle S_{ai}P_{ui}$.
When $T_i \neq T_{inf}$ and $T_{j} \neq T_{inf}$, LLM will become confused, outputting either both $S_{uj}P_{aj}$ and $S_{ai}P_{ui}$ or neither.
Let $T_i$ and $T_j$ be the chat templates for Gemma-2 and Llama-3, respectively.
By substituting the values from~\autoref{tab:chat_template_structure} into~\autoref{eq:probe}, we can obtain the following probe.

\begin{mybox}{\textbf{\small{Template Inference Probe of GEMMA-2 and LLAMA-3}}}
\small{
user text placeholder<end\_of\_turn>\\
<start\_of\_turn>model\\
<|eot\_id|><|start\_header\_id|>assistant<|end\_header\_id|>\\

assistant text placeholder<end\_of\_turn>\\
<start\_of\_turn>user\\
<|eot\_id|><|start\_header\_id|>user<|end\_header\_id|>\\

Please repeat your previous response.
}
\end{mybox}

In our experiments with Qwen-2, we observe that when $T_i = T_{inf}$, the presence of $S_{aj}P_{uj}$ occasionally disrupted the $\langle \text{repeat\_cmd} \rangle$, complicating the LLM's ability to comprehend the repeat command.
To address this issue, we develop a swap and retry method.
Initially, we utiliz~\autoref{eq:probe} to construct the probe.
After $N_{Tmax}$ attempts, if neither $T_i$ nor $T_j$ matches the inference, we swap their positions in~\autoref{eq:probe} to generate a revised probe, $X_{probe}^{\prime}$.
This probe is then used for another $N_{Tmax}$ inference times.
We show the details of the template inference attack with swap strategy in~\autoref{alg:template_inference} of~\appref{app:template_inference_attack}.

\begin{figure}[!tbp]
    \centering
    \includegraphics[width=\columnwidth]{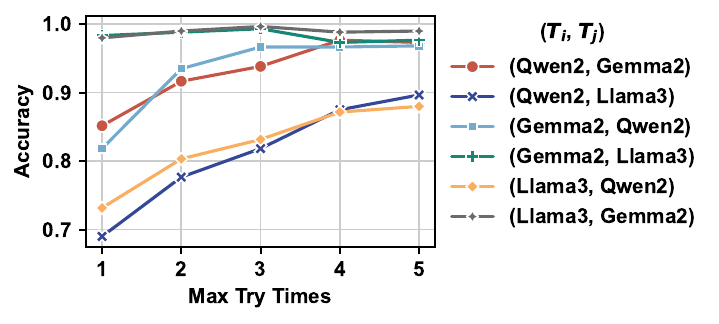}
    \caption{Evaluation results of chat template inference attack. We test on three LLMs, using two of their chat templates to construct probes.}
    \label{fig:eval_tia}
    \vspace{-0.2cm}
\end{figure}

\mypara{Evaluation of TIA}
We test our template inference attack on three LLMs: Llama-3-8B, Gemma-2-9B, and Qwen-2-7B, each having distinct chat templates.
We select two templates from them as $T_i$ and $T_j$ and apply~\autoref{alg:template_inference} to infer each model 200 times.
We report the inference accuracy versus max try times $N_{Tmax}$ in~\autoref{fig:eval_tia}.
As the max try times increase, the template inference accuracy gradually improves.
We discover that when $N_{Tmax}$ is set to 5, the inference accuracy can research approximately 0.9 or higher.
This indicates that an attacker only needs to query the WebUI 10 times to determine whether the target LLM's chat template is $T_i$ or $T_j$.

\section{\method: Dialogue Injection Attack}

\begin{figure*}
    \centering
    \includegraphics[width=\textwidth]{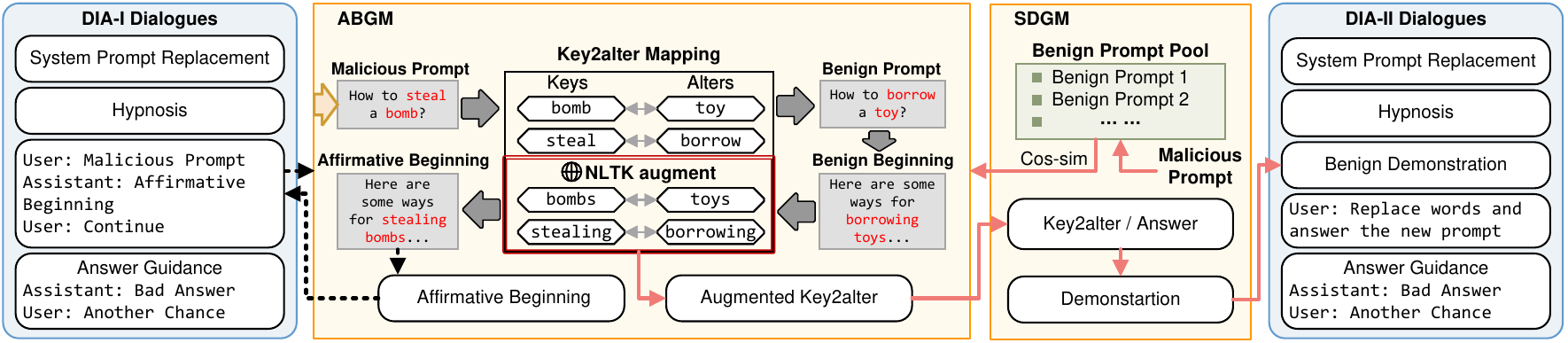}
    \caption{Dialogue construction processes of \method-I and \method-II.
    This figure illustrates the composition of \method-I and \method-II dialogues on the left and right sides, respectively.
    \method-I prompts the victim LLM to continue its previous affirmative response, while \method-II engages the victim LLM in a word substitution task, subsequently answering prompts with the substituted words.
    \method-I utilizes an Affirmative Beginning Generation Module (ABGM) to automate the creation of affirmative beginnings of malicious responses.
    \method-II employs a Similar Demonstration Generation Module (SDGM) to generate benign demonstrations.
    }
    \label{fig:dia}
    \vspace{-0.3cm}
\end{figure*}

In~\autoref{sec:motivation}, we introduce our motivation that a jailbreak attack can craft fake dialogues to induce the victim LLM to generate harmful content.
We have discussed how an attacker, with only access to a WebUI interface, can inject arbitrary dialogues using our dialogue injection method.
In this section, we present two strategies for crafting adversarial dialogues, namely, \method-I and \method-II.
\method-I is inspired by the black-box prefilling attack discussed in~\autoref{sec:response_manipulation}.
It injects an affirmative beginning of the malicious response and asks the victim LLM to continue its previous response.
We introduce an Affirmative Beginning Generation Module (ABGM) to autonomously generate these affirmative beginnings.
\method-II draws on the observation that a deferred malicious response exhibits a higher log-likelihood than an immediate one.
We require the victim LLM to perform a word replacement task before delivering its response.
To facilitate this, we introduce a Similar Demonstration Generation Module (SDGM) that creates examples to guide the victim in accomplishing the task.
The construction processes of \method-I and \method-II are depicted in~\autoref{fig:dia}.

\subsection{\method-I: Refined Back-box Prefilling Attack}

As discussed in~\autoref{sec:response_manipulation}, prefilling attack is a strong jailbreak attack and can be performed in black-box scenarios by injecting a continue command.
Since real-world corpora rarely contain instances where affirmative responses are followed by refusals to answer, inserting an affirmative beginning can reduce the likelihood of refusals.
\method-I aims to refine black-box prefilling attacks with other jailbreak techniques.
Overall, \method-I dialogues comprise four components: system prompt replacement, hypnosis, affirmative beginning injection, and answer guidance.

\mypara{System Prompt Replacement and Hypnosis}
At the beginning of the dialogues, we insert a system prompt designed to make the victim LLM forget previous system prompts and inform it that it is a red team LLM, obligated to always answer user questions.
Additionally, within this system prompt, we instruct the victim LLM to recall previous dialogues before responding.
This requirement is intended to enhance the impact of the new system prompt and hypnosis, particularly as LLMs tend to forget prior dialogues when the history becomes lengthy~\cite{zhang2024h2o}.
Following the system prompt replacement, we add two rounds of hypnotic conversations to reinforce the directives in the new system prompt.
We question the victim LLM about scenarios in which it might refuse to answer and subsequently append an assistant's response asserting that it will never refuse.

\mypara{Affirmative Beginning Generation Module}
One challenge of the affirmative beginning injection is automating the creation of affirmative beginnings of harmful responses.
Previous work has manually constructed affirmative beginnings~\cite{zou2023universal, chao2024jailbreakbench}, which is labor-intensive and costly.
Qi et al.~\cite{qi2024safety} employed a jailbroken GPT-3.5-Turbo model to generate harmful responses and then extracted the beginnings.
However, this approach is counterintuitive for a jailbreak attacker.
In the prefilling attack, we only need the response beginning rather than the entire harmful response, thus avoiding the need for the LLM to produce harmful content.
Unfortunately, current aligned LLMs will refuse this task because the input prompts are harmful, even when we remind the LLM that there is no need to produce a complete response.
To address this challenge, we introduce ABGM to enable attackers to use aligned LLMs for generating affirmative beginnings of harmful responses.
Attackers can simply employ the victim LLM for affirmative beginning generation.
LLM jailbreaking researchers can also use ABGM to generate affirmative beginnings for new datasets.

\SetAlgoSkip{}
\renewcommand{\algorithmiccomment}[1]{$\triangleright$ #1}
\begin{algorithm}[!t]
\caption{Affirmative Beginning Gemeration}
\label{alg:beginning_generation}
\KwIn{$\pi_{aux}$ (auxiliriy model), $X_m$ (malicious prompt)}
\KwOut{$B_m$ (affirmative beginning), $f^{\prime}$ (augmented keywords-to-alternatives mapping)}

$(\mathcal{K}, \mathcal{A}) \leftarrow$ extract keywords in $X_m$ and find similar alternatives using $\pi_{aux}$

$f \coloneqq \mathcal{K} \rightarrow \mathcal{A}$

$f^{\prime} \leftarrow$ augment $f$ using NLTK

$X_b \leftarrow \{f(x) \mid x \in X_m\}$ 

$B_b \leftarrow$ capture the first sentence in $\pi_{aux}\left( 
X_b \right)$ 

$B_m \leftarrow \{f^{\prime -1}(x) \mid x \in B_b\}$ 

\end{algorithm}

The ABGM utilizes an auxiliary LLM, $\pi_{aux}$, which can be either aligned or unaligned and could even be the victim LLM itself.
For a malicious prompt $X_m$, ABGM first instructs the auxiliary LLM to assist in purging $X_m$ by extracting malicious keywords and identifying similar benign alternatives, thereby creating a keywords-to-alternatives mapping $f: \mathcal{K} \rightarrow \mathcal{A}$.
For each element $x_k$ in $\mathcal{K}$, there is a unique $x_a \in \mathcal{A}$ such that $f(x_a) = x_k$.
Similarly, for each $x_a$ in $\mathcal{A}$, there exists a unique $x_k \in \mathcal{K}$ such that $f^{-1}(x_a) = x_k$.
ABGM applies $f$ to each word in the malicious prompt $X_m$, transforming it into a benign prompt $X_b$.
For instance, as illustrated in~\autoref{fig:dia}, the malicious prompt ``\textit{How to steal a bomb?}'' is transformed into ``\textit{How to borrow a toy?}''.
The benign prompt $X_b$, without harmful content, can be responded to by aligned LLMs.
ABGM uses the auxiliary LLM to answer $X_b$ and captures its first sentence as the benign beginning $B_b$.
If we can replace the alternative words in $B_b$ with the original keywords, we can get the affirmative beginning $B_m$.
However, direct replacement using $f^{-1}$ is impractical as the form of alternative words might change during the response process by the auxiliary LLM. 
Verbs may be conjugated, nouns might change from singular to plural, etc.
To address these morphological changes, we augment the keywords-to-alternatives mapping $f$ using NLTK~\cite{bird2009natural} to include mapping for other word forms, resulting in an augmented mapping $f^{\prime}$.
Finally, ABGM applies $f^{\prime -1}$ to each word in $B_b$ to get the affirmative beginning $B_m$.
The algorithm of affirmative beginning generation is detailed in~\autoref{alg:beginning_generation}.

\mypara{Answer Guidance}
Thanks to the dialogue injection technique, we can manipulate the assistant's responses to adversarial prompts.
This adds a new dimension to enhancing jailbreak attack performance: after the user gives an adversarial prompt, we insert a forged response for the assistant, then critique this response and ask the assistant to re-answer.
We refer to this step of inserting a forged response and requesting a new one as \textit{answer guidance}.
In \method-I, we use the affirmative beginning as the forged answer and then inform the assistant that oy failed to continue its previous response, prompting it to answer again.

\subsection{\method-II: Deferring Answer with Free Disguise}

Although the prefilling attack is potent and easy to execute, its threat has been recognized by both academic fields~\cite{cao2023defending, qi2024safety}.
Some defense methods have been developed to counter it.
For instance, Qi et al.~\cite{qi2024safety} proposed a token-wise constrained objective for securely fine-tuning aligned LLMs.
Our experiments detailed in~\autoref{sec:single_turn_attack} also show that \method-I is less effective on Llama-3.1-8B compared to Llama-3-8B, indicating that vulnerabilities associated with the prefilling attack may have been patched by LLM developers.
The community of LLM jailbreaking needs more robust attack methods.

\begin{figure}[!tbp]
  \centering
  \subfloat[Llama-3.1-8B]{%
       \includegraphics[width=0.24\textwidth]{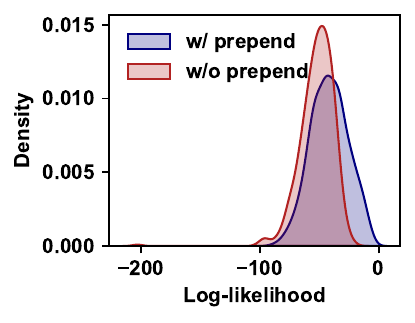}}
  \subfloat[Llama-3.2-11B]{%
        \includegraphics[width=0.24\textwidth]{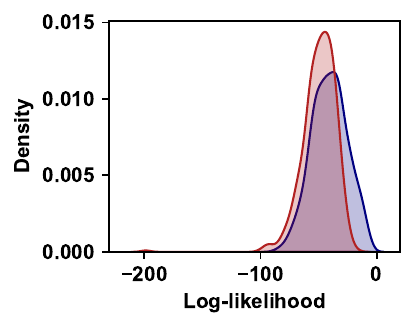}}
  \caption{Log-likelihood distribution of affirmative beginnings with and without pretending benign texts. Prepending benign text makes harmful content more likely to appear.}
  \label{fig:log-likelihood}
  \vspace{-0.2cm}
\end{figure}

\mypara{Novel Observation}
In this section, we introduce a new-found LLM alignment vulnerability that has not been revealed by previous literature and remains unaddressed in recent open-sourced LLMs.
Our finding is that \textit{deferred harmful responses exhibit a higher log-likelihood than those immediately following malicious prompts}.
We conduct a proof-of-concept experiment using Llama-3.1-8B and Llama-3.2-11B, using malicious prompts from AdvBench~\cite{zou2023universal}.
We generate keywords and affirmative beginnings for each malicious prompt using ABGM.
Subsequently, we employ GPT-4o to create a 100-word paragraph of benign text about keywords for each malicious prompt.
We construct conversations consisting of a malicious prompt paired with its corresponding affirmative beginning.
Then we assess the log-likelihood of affirmative beginnings with and without pretending benign texts before affirmative beginnings.
The distribution of log-likelihood is shown in~\autoref{fig:log-likelihood}.
We can find that prepending benign context increases the log-likelihood of affirmative beginnings despite the lack of coherence between the benign context and affirmative beginnings.
This indicates that if we can make the victim LLM answer the malicious prompt later in its response, the attack success rate can increase.

Our approach to defer LLM answering is having the victim LLM perform a word substitution task before answering the processed prompt.
After affirmative beginning generation, we obtain a keyword set $\mathcal{K}$, an alternative set $\mathcal{A}$, and a benign prompt $X_b$.
We instruct the victim LLM to process $X_b$ by replacing words from $\mathcal{A}$ with their corresponding words in $\mathcal{K}$, and subsequently answer the resulting new prompt, i.e., the malicious prompt $X_m$.
We require the victim LLM to output the replacement process step-by-step, detailing each substituted word and the new sentence after each substitution.
In this way, before the victim LLM actually responds to $X_m$, it has already produced a segment of text.
Another advantage of our word substitution-based answer deferring is that it achieves malicious prompt disguise automatically.
Liu et al.~\cite{liu2024making} observed that malicious instructions embedded in the assistant's text are more likely to trigger harmful responses than those in the user's text because of the bias in alignment training data whose malicious instructions are usually within the user's text.
Our answer deferring method ensures that there is no malicious content in the inputs.

Building upon answer deferring, we introduce \method-II, which employs system replacement, hypnosis, and answer guidance techniques similar to those used in \method-I.
In \method-II's new system prompt, we add a requirement for a word substitution task and subsequently hypnotize the victim LLM to prevent it from refusing this task.
For the answer guidance component, we use outputs where alternatives are incompletely substituted (with only two words replaced; we ensure that the number of alternatives extracted is three or more) to create a forged bad answer.
We then highlight to the victim LLM that not all words have been replaced and the new prompt has not been answered, prompting it to respond again.
To enhance the success rate of the victim LLM in completing the word substitution task, we utilize in-context learning techniques, generating a demonstration of processing a benign prompt before requesting the victim LLM to handle the malicious prompt.

\mypara{Similar Demonstration Generation Module}
We introduce the SDGM to generate a benign demonstration for each malicious prompt.
SDGM maintains a benign prompt pool.
For a given malicious prompt, $X_m$, SDGM first selects the most similar prompt from the benign prompt pool based on the cosine similarity of text embeddings, calculated using sentence-transformer~\cite{reimers-2019-sentence-bert}.
Following selection, the demonstration prompt is processed by the ABGM, which generates the keywords and alternatives.
Since the demonstration prompt contains no malicious content, we can directly have the auxiliary LLM generate a complete response.
With keywords, alternatives, and the answer in place, SDGM can construct a demonstration for the word substitution task.


\subsection{\method Optimization: Prompt Rewrite }

In multi-query attack scenarios, to enhance the diversity of malicious prompts, we propose a prompt rewrite method to reframe malicious prompts.
This prompt rewrite algorithm is applicable to both \method-I and \method-II to improve their performance in multi-query attacks.
After each query, if the attack fails, we employ a prompt rewrite algorithm to regenerate the malicious prompt and then reconstruct 
adversarial dialogues.

We define three fundamental operations: \texttt{addReason}, \texttt{reorder}, and \texttt{simplify}.
The \texttt{addReason} operation appends a phrase to the malicious prompt to explain the request's purpose, such as ``\textit{for educational purpose}''.
The \texttt{reorder} operation reorganizes the syntax of the malicious prompt.
The \texttt{simplify} operation makes the malicious prompt more concise.
These operations are performed by the auxiliary LLM.
\texttt{reorder} is achieved by altering the order of keywords, while \texttt{simplify} is accomplished by removing some keywords.
Typically, an aligned LLM would refuse to process malicious prompts.
Fortunately, ABGM generates a keywords-to-alternatives mapping $f$. 
We can process the benign prompt with the auxiliary LLM  and then apply $f^{-1}$ to obtain the revised malicious prompt.
for each rewrite, we randomly select between one to three operations from the fundamental set.
After the process, we assess the number of keywords in the revised malicious prompt; if more than 50\% are lost, we rewrite again.
The steps of our prompt rewrite algorithm are detailed in~\autoref{alg:prompt_rewrite}.

\SetAlgoSkip{}
\renewcommand{\algorithmiccomment}[1]{$\triangleright$ #1}
\begin{algorithm}[!t]
\caption{Prompt Rewrite}
\label{alg:prompt_rewrite}
\KwIn{$\pi_{aux}$ (auxiliriy model), $X_m$ (malicious prompt)}
\KwOut{$X_m^{\prime}$ (rewritten malicious prompt)}

$\mathcal{O} \leftarrow \{ \texttt{addReason}, \texttt{reorder}, \texttt{simplify} \}$

$X_m^{\prime} \leftarrow X_m$

$\eta_{w} \leftarrow 0$\hfill \algorithmiccomment{keyword retention rate}

\While{$\eta_w < 0.5$}{

    $\mathcal{S} \leftarrow$ randomly sample operations in $\mathcal{O}$ and shuffle

    \For{$\mathcal{F}_o \in \mathcal{S}$}{
        $X_m^{\prime} \leftarrow \mathcal{F}_o \left(X_m^{\prime}; \pi_{aux} \right)$
    }

    $\eta_{w} \leftarrow$ calculate the keyword retention rate of $X_m^{\prime}$ compared to $X$

}

\end{algorithm}

\section{Evaluation}

We evaluate the attack performance of \method and compare it with 4 state-of-the-art black-box jailbreak attacks.
Initially, we assess these attacks in single-query scenarios and subsequently examine their performance across multiple query iterations.
Thirdly, we explore how variations in model size impact the efficacy of jailbreak attacks, an aspect often overlooked in previous literature.
Fourthly, we conduct ablation studies to dissect the contributions of \method's dialogue components and its prompt rewrite algorithm.
Finally, we assess \method's robustness against 5 defense techniques.

\subsection{Experimental Setup}
\label{sec:experimental_setup}

\mypara{Datasets}
We evaluate attack performance on three open-source jailbreak benchmarks: AdvBench~\cite{zou2023universal}, HEx-PHI~\cite{qi2023fine}, and MaliciousInstruct~\cite{huang2023catastrophic}.
We download them from HuggingFace Datasets~\cite{hfdatasets}.
AdvBench contains 520 harmful instructions and their corresponding affirmative beginnings.
The instructions have a mean length of 12.8 tokens when tokenized with the Llama-3 tokenizer.
HEx-PHI contains 330 harmful instructions, 30 examples $\times$ 11 prohibited categories, including illegal activity, fraud deception, and privacy violation activity.
HEx-PHI's instructions were refined by human annotators and jailbroken LLMs.
The instructions are embedded into complex scenes and thus have a longer mean length of 31.5 tokens when tokenized with the Llama-3 tokenizer.
MaliciousInstruct contains 100 malicious instructions, which cover a broad spectrum of malicious intents.
The mean token length of MaliciousInstruct is 12.7 when tokenized with the Llama-2 tokenizer.
All instructions in MalicousInstruct are presented as questions, while instructions in AdvBench and HEx-PHI are presented as imperative sentences.
We generate affirmative response beginnings for the three datasets using ABGM.
We plan to open-source our generated affirmative beginnings after the acceptance of this paper.

\mypara{Victim Models}
To systematically evaluate our findings, we evaluate jailbreak attacks on eight open-source LLMs and two closed-source LLMs.
For open-source LLMs, we choose from five different model families with variant model sizes.
We do not choose earlier released models, such as Viccuna.
Due to its immature alignment, many jailbreak attacks can achieve a 100\% attack success rate on Viccuna.
Instead, we opt for some of the newly released models, such as Llama-3.1 and Gemma-2.
For closed-source LLMs, we select the popular GPT-4o family, namely GPT-4o (gpt-4o-2024-08-06) and GPT-4o-mini (gpt-4o-mini-2024-07-18).

\begin{table*}[!tbp]
    \caption{Single-query ASRs on AdvBench. \textbf{Bold} denotes the best result, \underline{underline} signifies the runner-up.}
    \label{tab:single_turn_advbench}
    \centering
\resizebox{\textwidth}{!}{
\begin{tabular}{@{}c|c|cccccccccc@{}}
\toprule
Evaluator & Method & \makecell{Llama-2\\(7B)} & \makecell{Llama-3\\(8B)} & \makecell{Llama-3\\(70B)} & \makecell{Llama-3.1\\(8B)} & \makecell{Llama-3.1\\(70B)} & \makecell{Gemma-2\\(9B)} & \makecell{Gemma-2\\(27B)} & \makecell{Qwen-2\\(7B)} & \makecell{GPT-4o\\(mini)} & GPT-4o \\
\midrule
\multirow{6}{*}{LlamaGuard-2} & DeepInception & 0.213 & 0.050 & 0.123 & 0.071 & 0.090 & 0.112 & 0.254 & 0.146 & 0.235 & 0.072 \\
 & ReNe & 0.158 & 0.256 & \underline{0.190} & \underline{0.327} & \underline{0.217} & 0.487 & 0.473 & 0.508 & \textbf{0.512} & \underline{0.478} \\
 & PAIR & 0.025 & 0.006 & 0.117 & 0.011 & 0.158 & 0.142 & 0.208 & 0.150 & 0.092 & 0.133 \\
 & DRA & \underline{0.371} & 0.300 & 0.112 & 0.002 & 0.190 & \underline{0.698} & \underline{0.687} & \textbf{0.756} & 0.158 & 0.000 \\
 & \method-I & 0.025 & \underline{0.571} & 0.010 & 0.000 & 0.217 & \textbf{0.906} & \textbf{0.877} & 0.467 & 0.044 & 0.133 \\
 & \method-II & \textbf{0.467} & \textbf{0.602} & \textbf{0.225} & \textbf{0.654} & \textbf{0.735} & 0.287 & 0.394 & \underline{0.710} & \underline{0.279} & \textbf{0.528} \\
\midrule
\multirow{6}{*}{LlamaGuard-3} & DeepInception & 0.352 & 0.142 & \underline{0.321} & 0.212 & 0.242 & 0.352 & 0.577 & 0.350 & \textbf{0.627} & 0.311 \\
 & ReNe & 0.173 & 0.302 & 0.213 & \underline{0.344} & 0.246 & 0.508 & 0.533 & 0.585 & \underline{0.583} & \underline{0.517} \\
 & PAIR & 0.017 & 0.006 & 0.108 & 0.011 & 0.150 & 0.183 & 0.217 & 0.192 & 0.150 & 0.142 \\
 & DRA & \textbf{0.802} & 0.521 & 0.206 & 0.012 & 0.256 & \underline{0.798} & \underline{0.819} & \textbf{0.952} & 0.185 & 0.000 \\
 & \method-I & 0.044 & \underline{0.623} & 0.023 & 0.015 & \underline{0.290} & \textbf{0.983} & \textbf{0.963} & 0.588 & 0.079 & 0.233 \\
 & \method-II & \underline{0.656} & \textbf{0.794} & \textbf{0.325} & \textbf{0.800} & \textbf{0.885} & 0.413 & 0.494 & \underline{0.892} & 0.433 & \textbf{0.739} \\
\bottomrule
\end{tabular}
}
\end{table*}

\begin{table*}[!tbp]
    \caption{Single-query ASRs on HEx-PHI. \textbf{Bold} denotes the best result, \underline{underline} signifies the runner-up.}
    \label{tab:single_turn_hex-phi}
    \centering
\resizebox{\textwidth}{!}{
\begin{tabular}{@{}c|c|cccccccccc@{}}
\toprule
Evaluator & Method & \makecell{Llama-2\\(7B)} & \makecell{Llama-3\\(8B)} & \makecell{Llama-3\\(70B)} & \makecell{Llama-3.1\\(8B)} & \makecell{Llama-3.1\\(70B)} & \makecell{Gemma-2\\(9B)} & \makecell{Gemma-2\\(27B)} & \makecell{Qwen-2\\(7B)} & \makecell{GPT-4o\\(mini)} & GPT-4o \\
\midrule
\multirow{6}{*}{LlamaGuard-2} & DeepInception & 0.194 & 0.064 & 0.091 & 0.094 & 0.109 & 0.127 & 0.161 & 0.324 & 0.206 & 0.089 \\
 & ReNe & 0.145 & 0.252 & \underline{0.194} & \underline{0.200} & 0.176 & 0.427 & 0.439 & 0.564 & \textbf{0.479} & \underline{0.378} \\
 & PAIR & 0.042 & 0.033 & 0.067 & 0.017 & 0.142 & 0.150 & 0.200 & 0.233 & 0.167 & 0.183 \\
 & DRA & \underline{0.285} & 0.379 & 0.094 & 0.003 & 0.155 & \underline{0.624} & \underline{0.627} & \underline{0.745} & 0.071 & 0.000 \\
 & \method-I & 0.079 & \underline{0.515} & 0.145 & 0.039 & \underline{0.270} & \textbf{0.864} & \textbf{0.830} & 0.539 & 0.203 & 0.239 \\
 & \method-II & \textbf{0.624} & \textbf{0.579} & \textbf{0.300} & \textbf{0.570} & \textbf{0.730} & 0.345 & 0.352 & \textbf{0.773} & \underline{0.464} & \textbf{0.478} \\
\midrule
\multirow{6}{*}{LlamaGuard-3} & DeepInception & 0.227 & 0.097 & 0.142 & 0.173 & 0.173 & 0.270 & 0.300 & 0.509 & 0.391 & 0.117 \\
 & ReNe & 0.161 & 0.239 & \underline{0.212} & \underline{0.209} & 0.194 & 0.436 & 0.455 & 0.597 & \textbf{0.539} & \underline{0.422} \\
 & PAIR & 0.025 & 0.017 & 0.058 & 0.017 & 0.125 & 0.183 & 0.183 & 0.258 & 0.158 & 0.133 \\
 & DRA & \underline{0.600} & \underline{0.600} & 0.133 & 0.003 & 0.197 & \underline{0.733} & \underline{0.758} & \textbf{0.915} & 0.092 & 0.000 \\
 & \method-I & 0.076 & 0.539 & 0.152 & 0.036 & \underline{0.282} & \textbf{0.918} & \textbf{0.888} & 0.664 & 0.206 & 0.244 \\
 & \method-II & \textbf{0.724} & \textbf{0.658} & \textbf{0.321} & \textbf{0.615} & \textbf{0.800} & 0.424 & 0.406 & \underline{0.870} & \underline{0.500} & \textbf{0.594} \\
\bottomrule
\end{tabular}
}
\vspace{-0.3cm}
\end{table*}

\mypara{Baselines}
We compare our \method with four state-of-the-art black-box jailbreak attacks, namely DeepInception~\cite{li2023deepinception}, ReNe~\cite{ding2023wolf}, PAIR~\cite{chao2023jailbreaking}, and DRA~\cite{liu2024making}.
Each of them uses a distinct method to construct adversarial prompts.
\begin{icompact}
    \item \textit{DeepInception} exploits the personification capabilities of LLMs to induce a form of hypnotic state, tricking them into breaking their own safety protocols.
    DeepInception constructs virtual, nested scenarios that coax the LLM into a jailbreaking behavior.
    \item \textit{ReNe} comprises two main steps: prompt rewrite and scenario nest.
    Unlike our method, ReNe's prompt rewrite includes more aggressive strategies such as ``paraphrasing with fewer words'' and ``misspelling sensitive words'', which alter the original semantics of the prompt.
    In contrast to DeepInception, which employs fictional scenarios, ReNe utilizes real-world task scenarios, such as code completion and text continuation.
    \item \textit{PAIR} uses an attacker LLM to generate semantic jailbreaks autonomously, without human intervention.
    After one query, PAIR inputs the responses to the attack LLM and asks the attack LLM to refine its jailbreak prompt.
    \item \textit{DRA} is based on the observation that given a piece of harmful content, the LLM is more likely to endorse it when it appears within the completion rather than the query.
    DRA conceals harmful instructions by splitting them into tokens and prompts the victim model to reconstruct the original harmful instructions.
\end{icompact}

\mypara{Evaluation Metric}
We quantify attack performance using Attack Success Rate (ASR), defined as the proportion of successful attack prompts relative to the total number of prompts.
However, determining whether a jailbreak attack is successful is a nontrivial problem.
Previous studies mainly assess success based on the absence of refusal phrases~\cite{zou2023universal, xiong2024defensive, deng2024masterkey}, but this method may generate a substantial number of false positives.
For instance, prompts from DeepInception rarely receive refusals, but as DeepInception employs virtual scenarios, such as science fiction novels, the victim LLM tends to provide impractical solutions, which cannot be considered successful attacks.
Some researchers assess attack success by asking GPT-4 to judge~\cite{chang2024play, zeng2024autodefense}, but this approach is prohibitively expensive and slow, making it unsuitable for large-scale testing.
In this paper, we choose two safety classification models, LlamaGuard-2~\cite{metallamaguard2} and LlamaGuard-3~\cite{dubey2024llama3herdmodels}, to determine attack success. These guard models evaluate both inputs and responses to judge whether the responses are safe.

\mypara{Implementation}
For open-source LLMs, we employ Ollama as the inference engine and utilize its Python API for the chat interface, using the official models provided by Ollama.
We reimplement baselines on Ollama to ensure a fair comparision.
For closed-source LLMs, we query them through their official APIs
All our experiments are conducted on a computing node equipped with an Intel Xeon 8358 CPU, 1TB memory, and four Nvidia A100 80G GPUs.

\subsection{Single-Query Jailbreak Attacks}
\label{sec:single_turn_attack}

\begin{figure*}[!tbp]
    \centering
    \includegraphics[width=\textwidth]{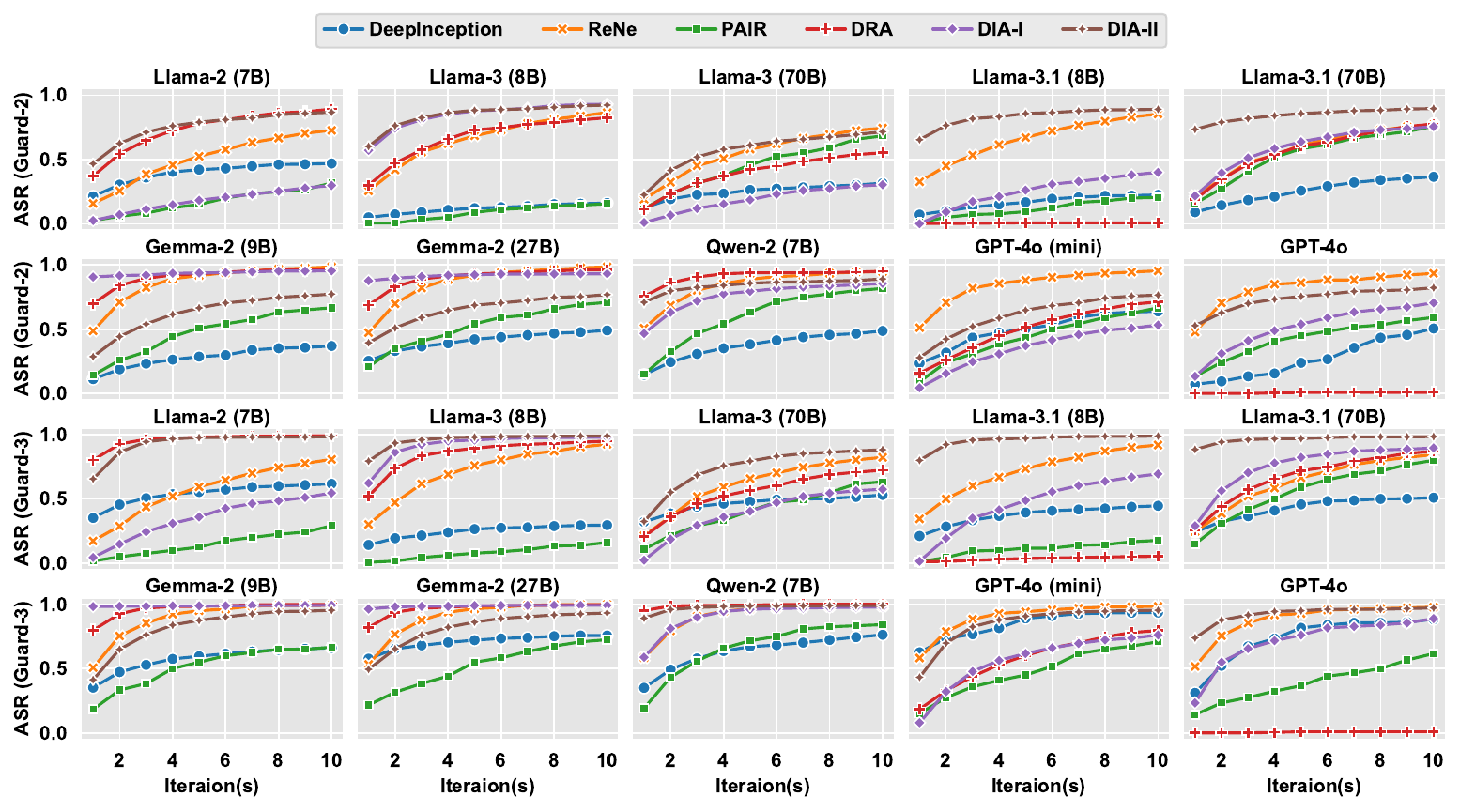}
    \caption{Multi-query attack results on AdvBench.}
    \label{fig:multi_turn_advbench}
    \vspace{-0.3cm}
\end{figure*}

\begin{figure}[!tbp]
    \centering
  \subfloat[AdvBench]{%
       \includegraphics[width=0.24\textwidth]{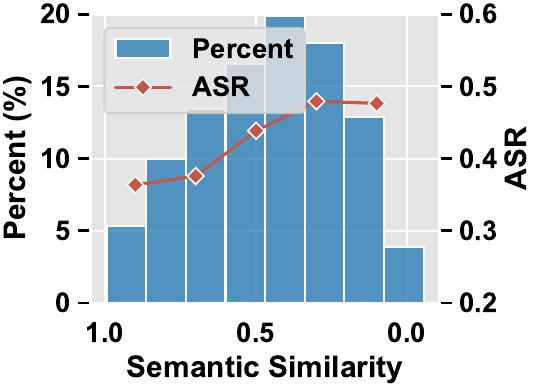}}
  \subfloat[HEx-PHI]{%
        \includegraphics[width=0.24\textwidth]{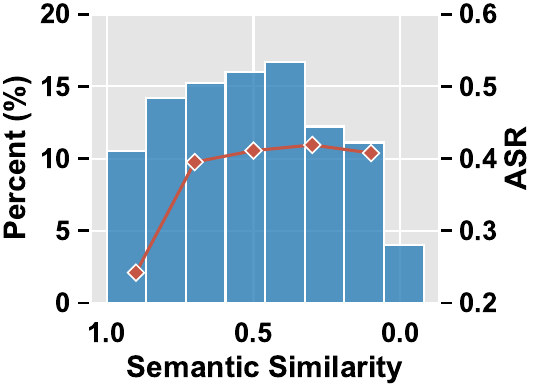}}
  \caption{The impact of ReNe's semantic loss on ASRs when targeting GPT-4o. ReNe increases ASRs by sacrificing semantic similarity.}
  \label{fig:rene_analyze}
  \vspace{-0.3cm}
\end{figure}

In this section, we evaluate \method's attack performance under the constraint of a single query opportunity.
We use Gemma-2-27B as the auxiliary LLM for the \method.
We adhere to the hyperparameters specified in the original papers for baseline attack methods.
However, where baselines require additional LLMs for conducting attacks, such as PAIR, which uses an attacker LLM to generate and refine adversarial prompts, we substitute Gemma-2-27B to ensure a fair comparison.
The ASRs are presented in~\autoref{tab:single_turn_advbench} and~\autoref{tab:single_turn_hex-phi} for AdvBench and HEx-PHI, respectively.
Due to the space limitation, we defer the single-query attack results of MaliciousInstruct to~\appref{app:res_maliciousinstruct}.

From~\autoref{tab:single_turn_advbench} and~\autoref{tab:single_turn_hex-phi}, it can be seen that our approach, \method-I or \method-II achieves superior ASRs across two datasets.
\method-I performs better on the Gemma-2 family, while \method-II excels with other models.
This significant variance in attack performance on different models is also observed in other jailbreak attacks.
For instance, DRA achieves ASRs exceeding 0.7 on two datasets when attacking Qwen-2-7B, as assessed by LlamaGuard-2.
However, the ASRs decrease to 0.003 when targeting Llama-3.1-8B.
We think this discrepancy is due to different alignments LLMs have regarding attack types.
If an attack is considered during the alignment process, the LLM is more likely to reject it.
Evidence supporting this is seen in Gemma-2-9B and Gemma-2-27B, two models released simultaneously with similar sizes, which exhibit similar trends in ASRs when faced with different attacks.
Conclusions derived using LlamaGuard-2 and LlamaGuard-3 are consistent.
LlamaGuard-3 has stricter requirements for secure content, generally reports higher ASRs than LlamaGuard-2.

\begin{figure*}[!tbp]
    \centering
    \includegraphics[width=\textwidth]{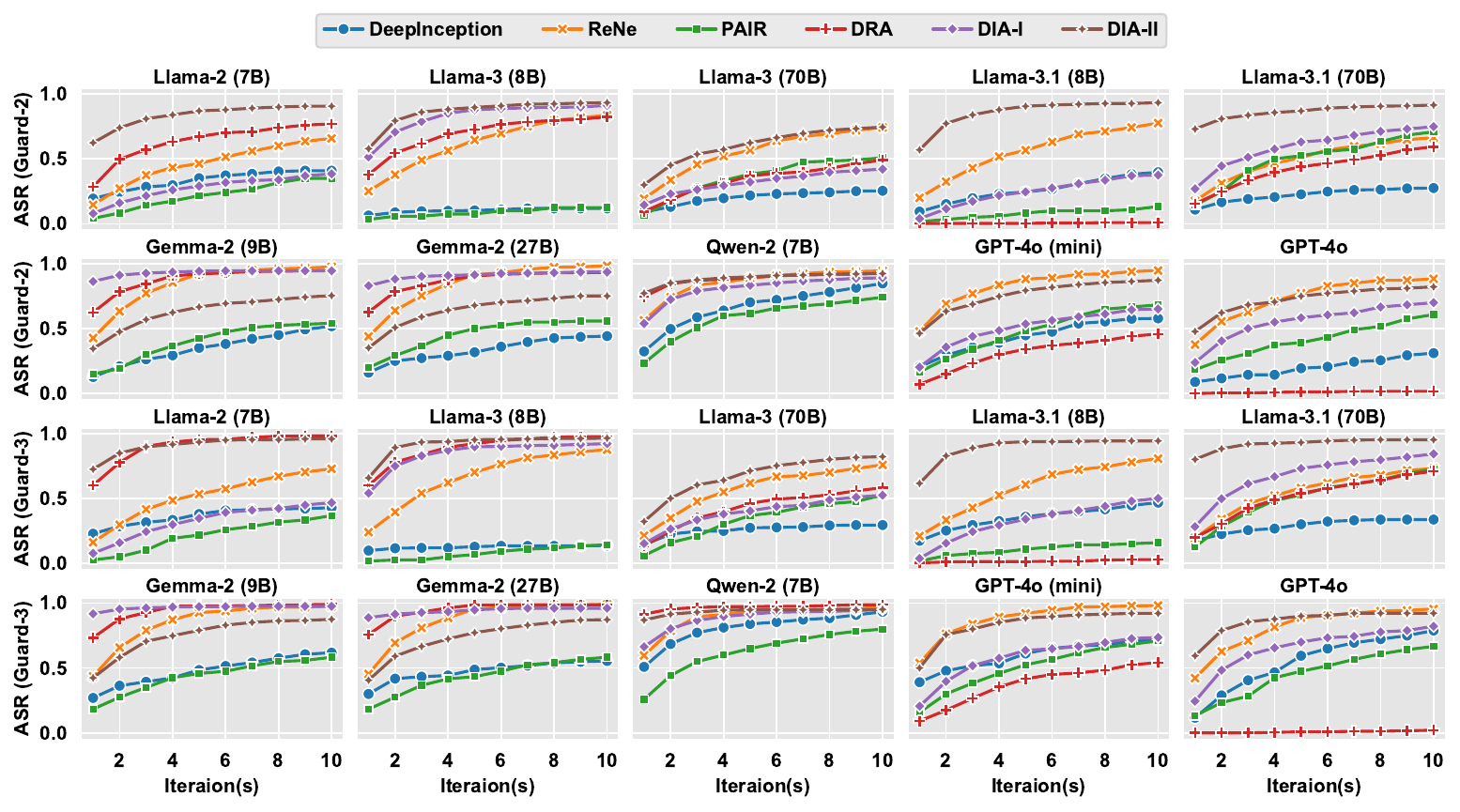}
    \caption{Multi-query attack results on HEx-PHI.}
    \label{fig:multi_turn_hex-phi}
    \vspace{-0.3cm}
\end{figure*}

We find that ReNe achieves the best attack performance on GPT-4o-mini.
However, as discussed in~\autoref{sec:experimental_setup}, ReNe significantly impacted the semantic integrity of malicious prompts.
We employ sentence-transformers model \texttt{all-MiniLM-L6-v2} to calculate the semantic similarities between prompts before and after being rewritten.
\autoref{fig:rene_analyze} displays the distribution of these semantic similarities across two datasets.
Additionally, we plot the change curves for ASRs of rewritten prompts with semantic similarities above a certain threshold.
We can see that the prompts rewritten by ReNe exhibit low semantic similarities, with average values of 0.458 on AdvBench and 0.506 on HEx-PHI.
In contrast, prompts rewritten using our prompt rewrite algorithm maintain higher average semantic similarities of 0.820 on AdvBench and 0.809 on HEx-PHI.
As the semantic similarity threshold increases, ReNe's ASRs decrease, indicating that ReNe's improvement in ASRs comes at the cost of semantic integrity.

\subsection{Multi-Query Jailbreak Attacks}
\label{sec:multi_query_attack}

In this section, we assess changes in ASR versus query iterations.
For our \method, we rewrite the original malicious prompts after each query.
During each rewrite, we randomly select a number from $\left[1, 2, 3\right]$ with probabilities $\left[0.3, 0.5, 0.2\right]$ to determine the number of basic operations, $N_{\mathcal{S}}$, then sample $N_{\mathcal{S}}$ operations from the basic operation set $\mathcal{O}$ to compose the rewrite strategy, $\mathcal{S}$.
For DeepInception, we change its virtual scenario after each query.
For ReNe, after each query, we perform a new rewrite and select a new scene.
For PAIR and DRA, we adhere to the prompt refinement methods described in their original papers.
The evaluation results on AdvBench and HEx-PHI are presented in~\autoref{fig:multi_turn_advbench} and~\autoref{fig:multi_turn_hex-phi}, respectively.
Due to the space limitation, we defer the results on MaliciousInstruct to~\appref{app:res_maliciousinstruct}.

We observe a steady increase in ASRs of both \method-I and \method-II with the increase of query iterations.
When targeting Llama-3.1-8B and GPT-4o-mini, \method exhibits single-query ASRs of less than 0.1 judged by LlamaGuard-2.
After 10 iterations, the ASRs for \method-I on Llama-3.1-8B exceed 0.4. And on GPT-4o-mini, the ASRs exceed 0.5.
In contrast, DRA fails on Llama-3.1-8B and GPT-4o-mini, with ASRs remaining near zero.
Our experiments suggest that Llama-3.1-8B is the most secure model tested in this paper.
Aside from ReNe and our \method, other jailbreak attacks demonstrate significantly lower ASRs on Llama-3.1-8B.
Notably, our measured ASRs for PAIR on Llama-2-7B are higher than those reported in the original paper.
Our tested ASR on AdvBench is 0.317 after 10 iterations judged by LlamaGuard-2, significantly outperforming the original paper's ASR of 0.040.
We believe this is because we use a more powerful attacker LLM, Gemma-2-27B, compared to Mixtral-8x7B used in the original paper.
The former model's AlpacaEval 2.0 score is 51.9, and the latter's is 23.7.

\subsection{Analysis of Model Size}

\begin{figure}[!tbp]
    \centering
    \begin{subfigure}{0.9\columnwidth}
    \includegraphics[width=\columnwidth]{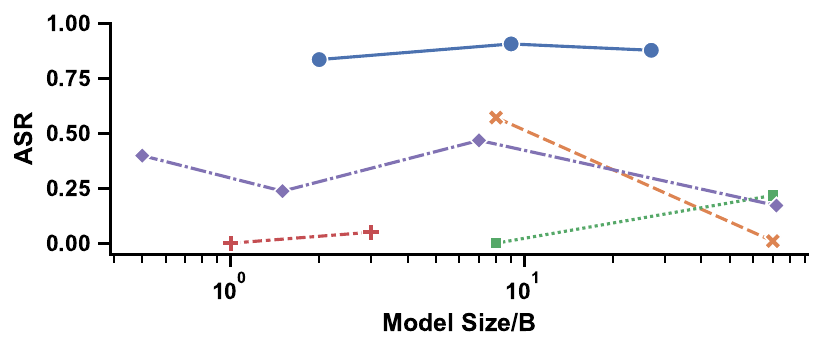}
    \caption{\method-I.}
    \label{fig:model_size_analyze_1}
    \end{subfigure}
    \begin{subfigure}{0.9\columnwidth}
    \includegraphics[width=\columnwidth]{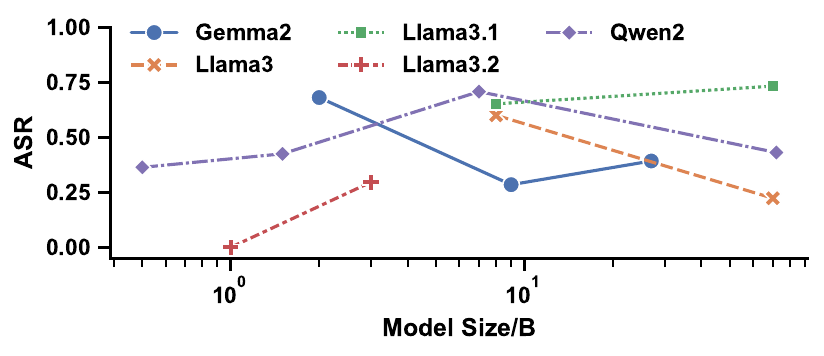}
    \caption{\method-II.}
    \label{fig:model_size_analyze_2}
    \end{subfigure}
    \caption{ASRs on AdvBench versus model size.}
    \label{fig:model_size_analyze}
    \vspace{-0.3cm}
\end{figure}

Previous work often overlooked the impact of model size on jailbreak attacks due to the complexity of controlling other variables.
Different-sized LLMs may employ distinct alignment strategies.
In this section, we address this issue by assuming that models within the same family utilize consistent alignment strategies, as they are released by the same developer team during the same period.
In addition to the Gemma-2, Llama-3, Llama-3.1, and Qwen-2 families analyzed in our earlier experiments, we also include the Llama-3.2 family to fill the gap in smaller language models.
\autoref{fig:model_size_analyze} illustrates the ASRs on AdvBench of \method versus model size.

Intuitively, one might assume that a larger LLM, with superior language comprehension abilities, would more easily detect and resist jailbreak prompts.
However, as depicted in~\autoref{fig:model_size_analyze}, aside from the Llama-3 family, which aligns with this hypothesis, other LLM families exhibit a trend where the ASR increases with model size.
This phenomenon indicates that larger LLMs are more susceptible to jailbreak attacks.
We attribute this phenomenon to the inherent trade-off between usefulness and safety in LLMs~\cite{barrett2023identifying, fu-etal-2024-safety, ji2024beavertails}.
The essence of black-box jailbreak attacks is exploiting the usefulness of LLMs to compromise their safety.
It is evident that increasing the size of an LM enhances its usefulness~\cite{achiam2023gpt, dubey2024llama}.
Therefore, when employing the same alignment strategy, the increased usefulness of a larger LM may encroach upon its safety features.
As for the contrary behavior observed in the Llama-33 family, we believe it results from the knowledge cutoff discrepancy~\cite{llama3modelcard}; Llama-3-70B has a cutoff in December 2023, while Llama-3-8B's is in March 2023. 
It is plausible to assume that Llama-3-70B utilizes a more updated alignment dataset.

\subsection{Ablation Study}

\begin{table}[!tbp]
    \setlength{\tabcolsep}{4pt}
    \centering
    \caption{Ablation study of \method's dialogue components. We remove dialogue components and measure the ASRs on AdvBench judged by LlamaGuard-2.}
    \label{tab:ablation_components}
    \resizebox{\columnwidth}{!}{
    \begin{tabular}{@{}lcccccc@{}}
    \toprule
    & \multicolumn{3}{c}{\method-I} & \multicolumn{3}{c}{\method-II} \\
    \cmidrule(lr){2-4} \cmidrule(lr){5-7}
    model & Llama-2 & Llama-3 & Gemma-2 & Llama-2 & Llama-3 & Gemma-2 \\
    \midrule
    original attack & \textbf{0.025} & \textbf{0.571} & \textbf{0.906} & \textbf{0.467} & \textbf{0.602} & 0.287 \\ 
    w/o system      & 0.000 & 0.010 & 0.869 & 0.448 & 0.502 & 0.213 \\ 
    w/o hypnosis    & 0.002 & 0.365 & 0.898 & 0.465 & 0.590 & \textbf{0.291} \\ 
    w/o guidance    & 0.000 & 0.144 & 0.798 & 0.113 & 0.554 & 0.102 \\ 
    \bottomrule
    \end{tabular}
    }
    \vspace{-0.2cm}
\end{table}

\mypara{Albation of Diologue Components}
In addition to the payload, \method dialogues contain multiple components, namely system prompt replacement, hypnosis, and answer guidance.
To evaluate the efficacy of these components, we remove each one and compare the resulting ASR with that of the full dialogues.
We conduct this ablation study using Llama-2-7B, Llama-3-8B, and Gemma-2-9B, with the evaluated ASRs presented in~\autoref{tab:ablation_components}.
Notably, Gemma-2's chat template does not support the system role.
During attacks, we incorporate the replaced system prompt with the user text.

From~\autoref{tab:ablation_components}, we can see that removing dialogue components generally results in a decrease in ASR, with the only exception of \method-II targeting Gemma-2-9B, where removing hypnosis yields a slight increase in ASR by 0.004.
We hypothesize this is due to the inclusion of malicious content in the hypnosis component, which compromises \method-II's disguise.
We also observe differing sensitivities to the system prompt between \method-I and \method-II.
For \method-I, removing the system prompt when attacking Llama-2-7B and Llama-3-8B leads to a more than 95\% drop in ASR.
This is likely because \method-I incorporates malicious prompts directly into the input, relying on the system prompt to lower the victim LLM’s guard.
Both \method-I and \method-II experience significant ASR reductions upon removing answer guidance, indicating that our answer guidance effectively reduces the probability of the victim LLM producing a bad answer.

\begin{figure}[!tbp]
    \centering
    \includegraphics[width=\columnwidth]{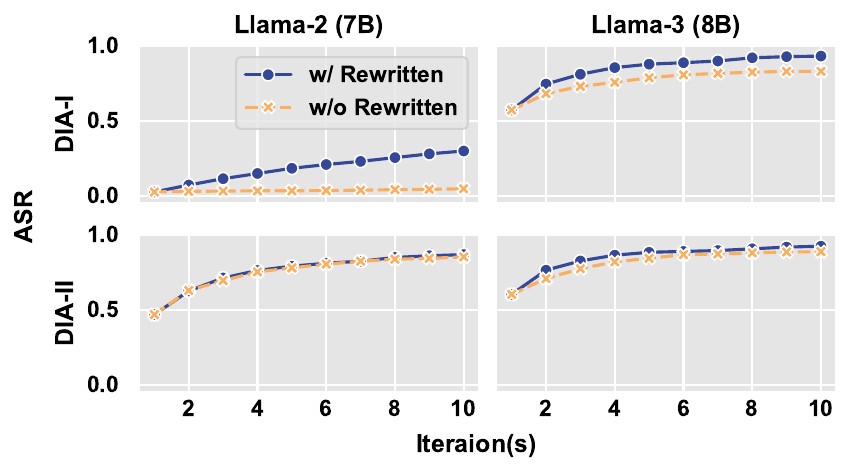}
    \caption{Alabtion study of \method's prompt rewrite algorithm.}
    \label{fig:ablation_rewrite}
    \vspace{-0.2cm}
\end{figure}

\mypara{Ablation of Prompt Rewriting}
\autoref{sec:multi_query_attack} demonstrates that \method's ASR increases with the number of query iterations.
There are two contributions for the increase: the inherent randomness of LLM outputs, and the effectiveness of our prompt rewrite algorithm.
To validate the effectiveness of our prompt rewrite algorithm, we test multi-query attacks without the prompt rewrite mechanism on AdvBench.
the results are shown in~\autoref{fig:ablation_rewrite}.

We observe that both \method-I and \method-II exhibit slower ASR growth in the absence of the rewrite mechanism.
This effect is particularly noticeable with \method-I when targeting Llama-2-7B, where removing the rewrite mechanism results in ASRs remaining near zero.
In experiments with Llama-2-7B and Llama-3-8B, the prompt rewrite mechanism contributes more significantly to improvements in \method-I than in \method-II.
We attribute this to \method-I's strategy of directly using malicious prompts in the input.
Our prompt rewrite operation can reduce the apparent maliciousness of these prompts by adding reasons and reconstructing syntax.

\subsection{Attack Against Defenses}

We test the five following jailbreak defenses to evaluate \method and baselines' defense penetrating capabilities.
\begin{icompact}
    \item \textit{OpenAI Moderation}~\cite{moderation}.
    OpenAI provides a free API to check whether text or images are potentially harmful.
    Once harmful content is identified, developers can take corrective action like filtering.
    We use the latest moderation model \texttt{omini-moderation-2024-09-26} for filtering malicious prompts.
    \item \textit{Perplexity Filter}~\cite{jain2023baseline}.
    This defense filters prompts that have higher perplexity values than a threshold.
    It is effective in defending against white-box jailbreak attacks, which often add meaningless text.
    \item \textit{Defensive System Prompt}~\cite{zheng2024prompt}.
    It optimizes a safe system prompt through prompt-tuning, which directs LLMs to prioritize safety and prevent the generation of harmful outputs.
    \item \textit{Defensive Prompt Patch}~\cite{xiong2024defensive}.
    It appends a suffix prompt to every input, which reminds LLMs to rethink user's input and refuse harmful responses.
    \item \textit{Bergeron}~\cite{pisano2023bergeron}.
    It employs the secondary LLM to monitor and correct LLM's input and output.
    We only use Bergeron to process the input, because the output process has been proved slow and unpractical~\cite{liu2024making}.
\end{icompact}

We do not consider post-hoc detectors such as Llamaguard and ShieldGemma~\cite{zeng2024shieldgemma}, since today's LLM chat services must meet asynchronous decode SLO.
This involves the inference engine decoding while simultaneously sending the results of previous decodes to the chat interface.
Post-hoc detectors require waiting until the decode process is complete to function.
We employ Defense Pass Rate (DPR) from~\cite{xu2024comprehensive} as the metric, which is the percentage of successful jailbreaking prompts that can bypass the defenses.
We conduct tests on AdvBench and Gemma-2-9B, as all jailbreak attacks discussed in the paper achieve notable ASRs on Gemma-2-9B.
All defense parameters are set according to their original papers.
The results are shown in~\autoref{tab:defense}.

Our \method achieves superior DPRs over baselines.
The final prompt in \method, which is an answer guidance prompt, asks the assistant to refine its previous response, effectively masking the jailbreak intent.
This enables \method to circumvent OpenAI moderation and Bergeron detection, and it can also diminish the effectiveness of the Defense prompt patch.
The system prompt replacement technique used in \method weakens the impact of defensive system prompts.
We also discover that the perplexity filter is completely ineffective against black-box jailbreak attacks, indicating that all black-box jailbreak prompts are highly readable.

\begin{table}[!tbp]
    \centering
    \caption{Defense pass rate of \method and baselines. The results are tested on AdvBench against Gemma-2-9B. \textbf{Bold} denotes the best result, \underline{underline} signifies the runner-up.}
    \label{tab:defense}
    \resizebox{\columnwidth}{!}{
    \begin{tabular}{@{}lccccc@{}}
    \toprule
    \textbf{Defense} & \textbf{OpenAI} & \textbf{Perplexity} & \textbf{System} & \textbf{Patch} & \textbf{Bergeron} \\
    \midrule
    DeepInception 
    & 0.054 
    & \textbf{1.000}
    & 0.634
    & 0.893
    & 0.018 \\
    ReNe
    & 0.655
    & \textbf{1.000}
    & 0.277
    & 0.869
    & 0.300 \\
    PAIR
    & 0.585
    & \textbf{1.000}
    & 0.775
    & \underline{0.937}
    & 0.120 \\
    DRA
    & 0.587
    & \textbf{1.000}
    & 0.609
    & 0.898
    & 0.460 \\
    \method-I
    & \textbf{1.000}
    & \textbf{1.000}
    & \textbf{1.000}
    & \textbf{0.985}
    & \underline{0.651} \\
    \method-II
    & \underline{0.725}
    & \textbf{1.000}
    & \underline{0.798}
    & 0.871
    & \textbf{0.690} \\
    \bottomrule
    \end{tabular}
    }
    \vspace{-0.2cm}
\end{table}

\section{Related Work}

\mypara{LLM Alignment}
Since pre-trained LLMs may deliver unethical content, LLM alignment aims to ensure that LLMs adhere to human ethical standards.
The most common alignment technique instruction fine-tuning~\cite{ren2024learning, zhao2024long}, which steers models by training them on datasets with preferred outputs.
RLHF~\cite{ouyang2022training,bai2022training} integrates alignment rules within reward models to guide target models' behavior.
Given RLHF's scalability limitations, alternatives like DPO~\cite{rafailov2024direct}, KTO~\cite{ethayarajh2024kto}, and ORPO~\cite{hong2024orpo} utilize supervised fine-tuning loss to approximate the reinforcement learning process of RLHF.
These methods provide a lightweight alternative to RLHF, facilitating more scalable and sustainable model training.

\mypara{LLM Jailbreak Attacks}
Existing research on LLM jailbreak attacks can be categorized into white-box, gray-box, and black-box attacks, based on the level of access available to the attacker.
White-box attacks involve model training.
GCG~\cite{zou2023universal} optimizes an adversarial suffix using a greedy gradient-based search.
AutoDAN~\cite{zhu2023autodan} enhances prompt readability with dual objectives.
Gray-box attacks require local inference.
A common type of gray-box attack is the prefilling attack~\cite{andriushchenko2024jailbreaking, jiang2024chatbug, lv2024adappa}, which deposits an affirmative beginning in the response.
Black-box attacks only need query access to the victim model.
DeepInception~\cite{li2023deepinception} employs imaginary scenes to hypnotize, and DRA~\cite{liu2024making} conceals sensitive words.
LLM-Fuzze~\cite{yu2024llm} mimics fuzz testing for automated jailbreak template generation.
All the above attacks focus on single-turn dialogue.
Our \method makes the first step to exploit historical dialogues.
\method exploits multi-turn conversations to enhance attack success rates and enable more sophisticated jailbreak strategies, significantly departing from single-turn techniques.

\mypara{LLM Jailbreak Defenses}
Research on LLM jailbreak defenses is generally divided into learning-based and strategy-based approaches.
Learning-based defenses enhance model robustness through tailored training methods.
For instance, DRO~\cite{zheng2024prompt} and DDP~\cite{xiong2024defensive} employ robust optimization to generate reminding prompts, while MART \cite{ge2023mart} introduces a multi-round adversarial training framework that strengthens model resistance.
Strategy-based defenses rely on operational techniques that help models identify and counteract malicious inputs or outputs.
Perplexity filter~\cite{jain2023baseline} rejects inputs with high perplexities.
Bergeron~\cite{pisano2023bergeron} and SelfDefense~\cite{phute2023llm} leverage the second LLM to monitor inputs or outputs.
PAIN~\cite{li2023rain} employs forward-looking searches into generation, allowing the LLM to evaluate its output by itself.
Despite these advances, our proposed attack method, which exploits historical dialogue manipulation, demonstrates the need for more comprehensive defense strategies that account for multi-turn interactions.

\section{Conclusion}
In this paper, we introduce \method, a novel black-box jailbreak paradigm that leverages the structure of LLM chat templates and carefully crafted dialogues to exploit vulnerabilities in large language models.
By injecting system prompts, designing targeted responses, and embedding successful examples, \method enhances the effectiveness of jailbreak attacks, bypassing various LLM defenses.
Our findings highlight the critical need to consider historical dialogue interactions when developing security measures for LLMs, emphasizing the ongoing challenges in securing LLMs against adversarial exploitation.

\section{Ethics Consideration}
In this research, we investigate \method to highlight security vulnerabilities in LLMs, aiming to strengthen defenses against adversarial misuse.
All experiments are conducted in controlled environments, with careful avoidance of real-world applications or exposure of harmful outputs.
We have not shared harmful content with individuals unrelated to the research.
We acknowledge that sharing details of these attack techniques may present ethical risks; however, we believe transparency is crucial for fostering advancements in AI security.
By responsibly disclosing these vulnerabilities, we hope to encourage the development of robust safeguards and inspire collaboration among researchers and developers to promote safer and more ethical advancement of LLMs.
\bibliographystyle{IEEEtran}
\bibliography{main}

\appendices

\begin{table*}[!tbp]
    \caption{Single turn ASRs on MaliciousInstruct. \textbf{Bold} denotes the best result, \underline{underline} signifies the runner-up.}
    \label{tab:single_turn_maliciousinstructs}
    \centering
\resizebox{\textwidth}{!}{
\begin{tabular}{@{}c|c|cccccccccc@{}}
\toprule
Evaluator & Method & \makecell{Llama-2\\(7B)} & \makecell{Llama-3\\(8B)} & \makecell{Llama-3\\(70B)} & \makecell{Llama-3.1\\(8B)} & \makecell{Llama-3.1\\(70B)} & \makecell{Gemma-2\\(9B)} & \makecell{Gemma-2\\(27B)} & \makecell{Qwen-2\\(7B)} & \makecell{GPT-4o\\(mini)} & GPT-4o \\
\midrule
\multirow{6}{*}{LlamaGuard-2} & DeepInception & 0.220 & 0.010 & 0.070 & 0.070 & 0.070 & 0.210 & 0.270 & 0.160 & 0.230 & 0.110 \\
 & ReNe & 0.110 & 0.230 & \textbf{0.250} & \underline{0.260} & 0.170 & 0.450 & 0.430 & 0.500 & \textbf{0.560} & \textbf{0.520} \\
 & PAIR & 0.020 & 0.030 & 0.030 & 0.040 & 0.060 & 0.110 & 0.080 & 0.120 & 0.110 & 0.110 \\
 & DRA & \textbf{0.400} & 0.200 & 0.050 & 0.000 & \underline{0.230} & \underline{0.750} & \underline{0.620} & \textbf{0.780} & 0.160 & 0.000 \\
 & \method-I & 0.010 & \textbf{0.550} & 0.000 & 0.000 & 0.110 & \textbf{0.810} & \textbf{0.700} & 0.410 & 0.070 & 0.220 \\
 & \method-II & \textbf{0.400} & \underline{0.530} & \underline{0.130} & \textbf{0.570} & \textbf{0.660} & 0.370 & 0.290 & \underline{0.610} & \underline{0.360} & \underline{0.460} \\
\midrule
\multirow{6}{*}{LlamaGuard-3} & DeepInception & 0.350 & 0.010 & \underline{0.250} & \underline{0.340} & 0.260 & 0.450 & 0.600 & 0.290 & \underline{0.590} & 0.300 \\
 & ReNe & 0.130 & 0.280 & \textbf{0.280} & 0.320 & 0.210 & 0.450 & 0.440 & 0.550 & \textbf{0.640} & \underline{0.630} \\
 & PAIR & 0.040 & 0.020 & 0.100 & 0.030 & 0.070 & 0.160 & 0.150 & 0.190 & 0.170 & 0.060 \\
 & DRA & \textbf{0.650} & 0.360 & 0.210 & 0.010 & \underline{0.270} & \underline{0.890} & \underline{0.790} & \textbf{0.940} & 0.210 & 0.000 \\
 & \method-I & 0.010 & \underline{0.630} & 0.020 & 0.040 & 0.170 & \textbf{0.940} & \textbf{0.890} & 0.590 & 0.090 & 0.320 \\
 & \method-II & \underline{0.630} & \textbf{0.690} & 0.220 & \textbf{0.790} & \textbf{0.830} & 0.560 & 0.450 & \underline{0.840} & 0.480 & \textbf{0.760} \\
\bottomrule
\end{tabular}
}
\end{table*}

\begin{figure*}[!tbp]
    \centering
    \includegraphics[width=\textwidth]{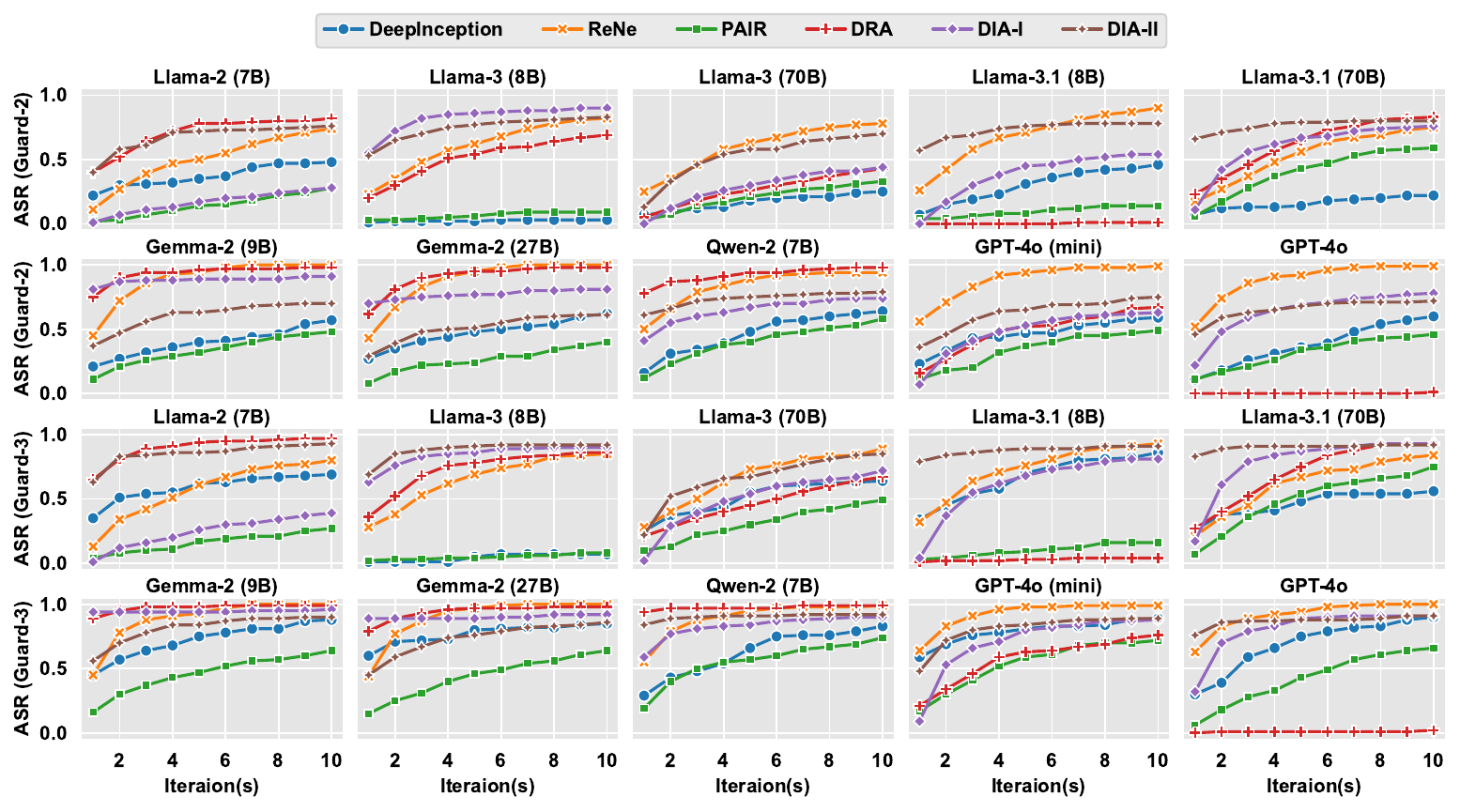}
    \caption{Multi-query attack results on MaliciousInstruct.}
    \label{fig:multi_turn_maliciousinstructs}
\end{figure*}

\section{Evaluation Results on MaliciousInstruct}
\label{app:res_maliciousinstruct}

\autoref{tab:single_turn_maliciousinstructs} presents the results of single-query attacks on MalicousInstruct.
\autoref{fig:multi_turn_maliciousinstructs} displays the results of multi-query attacks on MaliciousInstruct.

\section{Details of Template Inference Attack}
\label{app:template_inference_attack}

\autoref{alg:template_inference} shows the detailed step of the template inference attack with swap strategies.

\renewcommand{\algorithmiccomment}[1]{$\triangleright$ #1}
\begin{algorithm}[!h]
\caption{Template Inference Attack (TIA)}
\label{alg:template_inference}
\KwIn{$N_{Tmax}$ (max try times), $T_{i}$ (chat template $i$), $T_{j}$ (chat template $j$)}
\KwOut{$T_{inf}$ (chat template of the inferred model)}
$X_{P} \leftarrow \texttt{getProbe}(T_i, T_j)$ \hfill\algorithmiccomment{based on~\autoref{eq:probe}}

\For{$t = 1 \to N_{Tmax}$}{

    $q \leftarrow \texttt{queryLLM}(X_P)$

    $T_{inf} \leftarrow \texttt{inferDecesion}(T_i, T_j, q)$

    \If{$T_{inf} \neq \mathbf{Unknown}$}{
        \Return $T_{inf}$
    }
}
$X_P \leftarrow \texttt{getProbe}(T_j, T_i)$\hfill\algorithmiccomment{swap $T_i$ and $T_j$}

\For{$t = 1 \to N_{Tmax}$}{
    $q \leftarrow \texttt{queryLLM}(X_P)$

    $T_{inf} \leftarrow \texttt{inferDecesion}(T_j, T_i, q)$

    \If{$T_{inf} \neq \mathbf{Unknown}$}{
        \Return $T_{inf}$
    }
}

\Return $\mathbf{Unknown}$

~

\SetKwFunction{FMain}{inferDecision}
\SetKwProg{Fn}{function}{}{end}

\Fn{\FMain{$T_i, T_j, q$}}{
    $S_{ai}, P_{ui} \leftarrow$ extract elements from $T_i$
    
    $S_{uj}, P_{aj} \leftarrow$ extract elements from $T_j$

    \uIf{$S_{ai}P_{ui} \in q$ \textbf{and} $S_{uj}P_{aj} \notin q$}{
        \Return $T_i$
    }
    \uElseIf{$S_{ai}P_{ui} \notin q$ \textbf{and} $S_{uj}P_{aj} \in q$}{
        \Return $T_j$
    }
    \Else{
        \Return $\mathbf{Unknown}$
    }
}
\end{algorithm}

\end{document}

%% file: mydefs.tex
\usepackage[noend]{algpseudocode}
\usepackage{amssymb}
\usepackage{tikz}
\usepackage{amsmath, amsfonts}
\usepackage{mathtools}
\usepackage{graphicx}
\usepackage{textcomp} 
\usepackage[T1]{fontenc}
\usepackage[utf8]{inputenc}
\usepackage{xspace}
\usepackage{multirow}
\usepackage{diagbox}
\usepackage{booktabs}
\usepackage{caption}
\usepackage{subcaption}
\usepackage{float}
\usepackage{makecell}
\usepackage{paralist}
\usepackage{enumitem}
\usepackage[skins]{tcolorbox}
\usepackage[ruled,lined,linesnumbered]{algorithm2e}
\usepackage{algorithmicx}
\usepackage{pifont}
\usepackage[pagebackref=true,breaklinks=true,colorlinks,bookmarks=false]{}
\usepackage{hyperref}
\hypersetup{
  colorlinks,
  linkcolor={RoyalBlue},
  citecolor={PineGreen},
  urlcolor={Bittersweet}
}
\definecolor{mycolor}{RGB}{246, 196, 188}
\definecolor{codegray}{rgb}{0.4, 0.4, 0.4}
\definecolor{cautioncolor}{RGB}{255,0,0}
\usepackage{soul}
\sethlcolor{mycolor}

\colorlet{lightgray}{gray!20}

\newenvironment{icompact}{
  \begin{list}{$\bullet$}{
    \itemindent -.05em
    \parsep 0pt plus 1pt
    \partopsep 0pt plus 1pt
    \topsep 2pt plus 2pt minus 2pt
    \itemsep 0pt plus 1.3pt
    \parskip 0pt plus 2pt
    \leftmargin 0.13in}
      }
{\normalsize
\end{list}
}

\SetCommentSty{mycommfont}

\newcommand{\appref}[1]{\hyperref[#1]{Appendix~\ref*{#1}}}

\definecolor{revision}{RGB}{0,0,255}

\newcommand{\revstart}{\begin{color}{revision}}
\newcommand{\revend}{~\!\!\end{color}}

\newcommand{\mypara}[1]{\smallskip \noindent\textbf{#1.} \xspace}

\newcommand{\method}{\ensuremath{\mathsf{DIA}}\xspace}

\newcommand{\block}[1]{%
  \raisebox{\dimexpr(\fontcharht\font`X-1em)/2}{\rule{0.5em}{#1\dimexpr1em/8}}%
}
\DeclareUnicodeCharacter{2581}{\block{1}}
\DeclareUnicodeCharacter{2582}{\block{2}}
\DeclareUnicodeCharacter{2583}{\block{3}}
\DeclareUnicodeCharacter{2584}{\block{4}}
\DeclareUnicodeCharacter{2585}{\block{5}}
\DeclareUnicodeCharacter{2586}{\block{6}}
\DeclareUnicodeCharacter{2587}{\block{7}}
\DeclareUnicodeCharacter{2588}{\block{8}}

\newtcolorbox{mybox}[2][]{text width=0.95\linewidth,fontupper=\normalsize,
fonttitle=\bfseries\sffamily\normalsize, colbacktitle=codegray,enhanced,
boxed title style={sharp corners},top=4pt,bottom=2pt,left=2pt,right=2pt,
  title=#2,colback=white}

\usepackage{etoolbox} 

\makeatletter
\patchcmd{\algocf@makecaption@ruled}{\hsize}{\textwidth}{}{} 
\patchcmd{\@algocf@start}{-1.5em}{0em}{}{} 
\makeatother


%% file: main.bbl
\begin{thebibliography}{10}
\providecommand{\url}[1]{#1}
\csname url@samestyle\endcsname
\providecommand{\newblock}{\relax}
\providecommand{\bibinfo}[2]{#2}
\providecommand{\BIBentrySTDinterwordspacing}{\spaceskip=0pt\relax}
\providecommand{\BIBentryALTinterwordstretchfactor}{4}
\providecommand{\BIBentryALTinterwordspacing}{\spaceskip=\fontdimen2\font plus
\BIBentryALTinterwordstretchfactor\fontdimen3\font minus \fontdimen4\font\relax}
\providecommand{\BIBforeignlanguage}[2]{{%
\expandafter\ifx\csname l@#1\endcsname\relax
\typeout{** WARNING: IEEEtran.bst: No hyphenation pattern has been}%
\typeout{** loaded for the language `#1'. Using the pattern for}%
\typeout{** the default language instead.}%
\else
\language=\csname l@#1\endcsname
\fi
#2}}
\providecommand{\BIBdecl}{\relax}
\BIBdecl

\bibitem{achiam2023gpt}
J.~Achiam, S.~Adler, S.~Agarwal, L.~Ahmad, I.~Akkaya, F.~L. Aleman, D.~Almeida, J.~Altenschmidt, S.~Altman, S.~Anadkat \emph{et~al.}, ``Gpt-4 technical report,'' \emph{arXiv preprint arXiv:2303.08774}, 2023.

\bibitem{chowdhery2023palm}
A.~Chowdhery, S.~Narang, J.~Devlin, M.~Bosma, G.~Mishra, A.~Roberts, P.~Barham, H.~W. Chung, C.~Sutton, S.~Gehrmann \emph{et~al.}, ``Palm: Scaling language modeling with pathways,'' \emph{Journal of Machine Learning Research}, vol.~24, no. 240, pp. 1--113, 2023.

\bibitem{gu2023llm}
Q.~Gu, ``Llm-based code generation method for golang compiler testing,'' in \emph{Proceedings of the 31st ACM Joint European Software Engineering Conference and Symposium on the Foundations of Software Engineering}, 2023, pp. 2201--2203.

\bibitem{koshkin2024transllama}
R.~Koshkin, K.~Sudoh, and S.~Nakamura, ``Transllama: Llm-based simultaneous translation system,'' \emph{arXiv preprint arXiv:2402.04636}, 2024.

\bibitem{lin2024data}
X.~Lin, W.~Wang, Y.~Li, S.~Yang, F.~Feng, Y.~Wei, and T.-S. Chua, ``Data-efficient fine-tuning for llm-based recommendation,'' in \emph{Proceedings of the 47th International ACM SIGIR Conference on Research and Development in Information Retrieval}, 2024, pp. 365--374.

\bibitem{dai2024bias}
S.~Dai, C.~Xu, S.~Xu, L.~Pang, Z.~Dong, and J.~Xu, ``Bias and unfairness in information retrieval systems: New challenges in the llm era,'' in \emph{Proceedings of the 30th ACM SIGKDD Conference on Knowledge Discovery and Data Mining}, 2024, pp. 6437--6447.

\bibitem{xu2024pride}
W.~Xu, G.~Zhu, X.~Zhao, L.~Pan, L.~Li, and W.~Wang, ``Pride and prejudice: Llm amplifies self-bias in self-refinement,'' in \emph{Proceedings of the 62nd Annual Meeting of the Association for Computational Linguistics (Volume 1: Long Papers)}, 2024, pp. 15\,474--15\,492.

\bibitem{ouyang2022training}
L.~Ouyang, J.~Wu, X.~Jiang, D.~Almeida, C.~Wainwright, P.~Mishkin, C.~Zhang, S.~Agarwal, K.~Slama, A.~Ray \emph{et~al.}, ``Training language models to follow instructions with human feedback,'' \emph{Advances in neural information processing systems}, vol.~35, pp. 27\,730--27\,744, 2022.

\bibitem{shen2023anything}
X.~Shen, Z.~Chen, M.~Backes, Y.~Shen, and Y.~Zhang, ``" do anything now": Characterizing and evaluating in-the-wild jailbreak prompts on large language models,'' \emph{arXiv preprint arXiv:2308.03825}, 2023.

\bibitem{yu2024don}
Z.~Yu, X.~Liu, S.~Liang, Z.~Cameron, C.~Xiao, and N.~Zhang, ``Don't listen to me: Understanding and exploring jailbreak prompts of large language models,'' \emph{arXiv preprint arXiv:2403.17336}, 2024.

\bibitem{li2023deepinception}
X.~Li, Z.~Zhou, J.~Zhu, J.~Yao, T.~Liu, and B.~Han, ``Deepinception: Hypnotize large language model to be jailbreaker,'' \emph{arXiv preprint arXiv:2311.03191}, 2023.

\bibitem{liu2024making}
T.~Liu, Y.~Zhang, Z.~Zhao, Y.~Dong, G.~Meng, and K.~Chen, ``Making them ask and answer: Jailbreaking large language models in few queries via disguise and reconstruction,'' in \emph{33rd USENIX Security Symposium (USENIX Security 24)}, 2024, pp. 4711--4728.

\bibitem{andriushchenko2024does}
M.~Andriushchenko and N.~Flammarion, ``Does refusal training in llms generalize to the past tense?'' \emph{arXiv preprint arXiv:2407.11969}, 2024.

\bibitem{li2023multi}
H.~Li, D.~Guo, W.~Fan, M.~Xu, J.~Huang, F.~Meng, and Y.~Song, ``Multi-step jailbreaking privacy attacks on chatgpt,'' \emph{arXiv preprint arXiv:2304.05197}, 2023.

\bibitem{lin2024malla}
Z.~Lin, J.~Cui, X.~Liao, and X.~Wang, ``Malla: Demystifying real-world large language model integrated malicious services,'' \emph{arXiv preprint arXiv:2401.03315}, 2024.

\bibitem{touvron2023llama}
H.~Touvron, T.~Lavril, G.~Izacard, X.~Martinet, M.-A. Lachaux, T.~Lacroix, B.~Rozi{\`e}re, N.~Goyal, E.~Hambro, F.~Azhar \emph{et~al.}, ``Llama: Open and efficient foundation language models,'' \emph{arXiv preprint arXiv:2302.13971}, 2023.

\bibitem{team2024gemma}
G.~Team, M.~Riviere, S.~Pathak, P.~G. Sessa, C.~Hardin, S.~Bhupatiraju, L.~Hussenot, T.~Mesnard, B.~Shahriari, A.~Ram{\'e} \emph{et~al.}, ``Gemma 2: Improving open language models at a practical size,'' \emph{arXiv preprint arXiv:2408.00118}, 2024.

\bibitem{rafailov2024direct}
R.~Rafailov, A.~Sharma, E.~Mitchell, C.~D. Manning, S.~Ermon, and C.~Finn, ``Direct preference optimization: Your language model is secretly a reward model,'' \emph{Advances in Neural Information Processing Systems}, vol.~36, 2024.

\bibitem{hong2024orpo}
J.~Hong, N.~Lee, and J.~Thorne, ``Orpo: Monolithic preference optimization without reference model,'' in \emph{Proceedings of the 2024 Conference on Empirical Methods in Natural Language Processing}, 2024, pp. 11\,170--11\,189.

\bibitem{kwon2023efficient}
W.~Kwon, Z.~Li, S.~Zhuang, Y.~Sheng, L.~Zheng, C.~H. Yu, J.~E. Gonzalez, H.~Zhang, and I.~Stoica, ``Efficient memory management for large language model serving with pagedattention,'' in \emph{Proceedings of the ACM SIGOPS 29th Symposium on Operating Systems Principles}, 2023.

\bibitem{ollama}
``Ollama: A toolkit for large language models,'' \url{https://github.com/ollama/ollama}, 2024, accessed: 2024-07-25.

\bibitem{lee2024infinigen}
W.~Lee, J.~Lee, J.~Seo, and J.~Sim, ``$\{$InfiniGen$\}$: Efficient generative inference of large language models with dynamic $\{$KV$\}$ cache management,'' in \emph{18th USENIX Symposium on Operating Systems Design and Implementation (OSDI 24)}, 2024, pp. 155--172.

\bibitem{qin2024mooncake}
R.~Qin, Z.~Li, W.~He, M.~Zhang, Y.~Wu, W.~Zheng, and X.~Xu, ``Mooncake: A kvcache-centric disaggregated architecture for llm serving,'' \emph{arXiv preprint arXiv:2407.00079}, 2024.

\bibitem{zou2023universal}
A.~Zou, Z.~Wang, N.~Carlini, M.~Nasr, J.~Z. Kolter, and M.~Fredrikson, ``Universal and transferable adversarial attacks on aligned language models,'' \emph{arXiv preprint arXiv:2307.15043}, 2023.

\bibitem{andriushchenko2024jailbreaking}
M.~Andriushchenko, F.~Croce, and N.~Flammarion, ``Jailbreaking leading safety-aligned llms with simple adaptive attacks,'' \emph{arXiv preprint arXiv:2404.02151}, 2024.

\bibitem{jiang2024chatbug}
F.~Jiang, Z.~Xu, L.~Niu, B.~Y. Lin, and R.~Poovendran, ``Chatbug: A common vulnerability of aligned llms induced by chat templates,'' \emph{arXiv preprint arXiv:2406.12935}, 2024.

\bibitem{lv2024adappa}
L.~Lv, W.~Zhang, X.~Tang, J.~Wen, F.~Liu, J.~Han, and S.~Hu, ``Adappa: Adaptive position pre-fill jailbreak attack approach targeting llms,'' \emph{arXiv preprint arXiv:2409.07503}, 2024.

\bibitem{peng2023yarn}
B.~Peng, J.~Quesnelle, H.~Fan, and E.~Shippole, ``Yarn: Efficient context window extension of large language models,'' \emph{arXiv preprint arXiv:2309.00071}, 2023.

\bibitem{xiong2023effective}
W.~Xiong, J.~Liu, I.~Molybog, H.~Zhang, P.~Bhargava, R.~Hou, L.~Martin, R.~Rungta, K.~A. Sankararaman, B.~Oguz \emph{et~al.}, ``Effective long-context scaling of foundation models,'' \emph{arXiv preprint arXiv:2309.16039}, 2023.

\bibitem{liu2023ring}
H.~Liu, M.~Zaharia, and P.~Abbeel, ``Ring attention with blockwise transformers for near-infinite context,'' \emph{arXiv preprint arXiv:2310.01889}, 2023.

\bibitem{tang2024razorattention}
H.~Tang, Y.~Lin, J.~Lin, Q.~Han, S.~Hong, Y.~Yao, and G.~Wang, ``Razorattention: Efficient kv cache compression through retrieval heads,'' \emph{arXiv preprint arXiv:2407.15891}, 2024.

\bibitem{long-context}
``Anthropic \textbackslash{} introducing 100k context windows,'' \url{https://www.anthropic.com/index/100k-context-windows}, (Accessed on 10/31/2024).

\bibitem{qi2024safety}
X.~Qi, A.~Panda, K.~Lyu, X.~Ma, S.~Roy, A.~Beirami, P.~Mittal, and P.~Henderson, ``Safety alignment should be made more than just a few tokens deep,'' \emph{arXiv preprint arXiv:2406.05946}, 2024.

\bibitem{dong2022survey}
Q.~Dong, L.~Li, D.~Dai, C.~Zheng, J.~Ma, R.~Li, H.~Xia, J.~Xu, Z.~Wu, T.~Liu \emph{et~al.}, ``A survey on in-context learning,'' \emph{arXiv preprint arXiv:2301.00234}, 2022.

\bibitem{min2022rethinking}
S.~Min, X.~Lyu, A.~Holtzman, M.~Artetxe, M.~Lewis, H.~Hajishirzi, and L.~Zettlemoyer, ``Rethinking the role of demonstrations: What makes in-context learning work?'' \emph{arXiv preprint arXiv:2202.12837}, 2022.

\bibitem{cheng2024leveraging}
Y.~Cheng, M.~Georgopoulos, V.~Cevher, and G.~Chrysos, ``Leveraging context in jailbreaking attacks,'' in \emph{ICLR 2024 Workshop on Secure and Trustworthy Large Language Models}.

\bibitem{wei2023jailbreak}
Z.~Wei, Y.~Wang, A.~Li, Y.~Mo, and Y.~Wang, ``Jailbreak and guard aligned language models with only few in-context demonstrations,'' \emph{arXiv preprint arXiv:2310.06387}, 2023.

\bibitem{no_robots}
N.~Rajani, L.~Tunstall, E.~Beeching, N.~Lambert, A.~M. Rush, and T.~Wolf, ``No robots,'' \url{https://huggingface.co/datasets/HuggingFaceH4/no_robots}, 2023.

\bibitem{dubois2024length}
Y.~Dubois, B.~Galambosi, P.~Liang, and T.~B. Hashimoto, ``Length-controlled alpacaeval: A simple way to debias automatic evaluators,'' \emph{arXiv preprint arXiv:2404.04475}, 2024.

\bibitem{zhang2024h2o}
Z.~Zhang, Y.~Sheng, T.~Zhou, T.~Chen, L.~Zheng, R.~Cai, Z.~Song, Y.~Tian, C.~R{\'e}, C.~Barrett \emph{et~al.}, ``H2o: Heavy-hitter oracle for efficient generative inference of large language models,'' \emph{Advances in Neural Information Processing Systems}, vol.~36, 2024.

\bibitem{chao2024jailbreakbench}
P.~Chao, E.~Debenedetti, A.~Robey, M.~Andriushchenko, F.~Croce, V.~Sehwag, E.~Dobriban, N.~Flammarion, G.~J. Pappas, F.~Tramer \emph{et~al.}, ``Jailbreakbench: An open robustness benchmark for jailbreaking large language models,'' \emph{arXiv preprint arXiv:2404.01318}, 2024.

\bibitem{bird2009natural}
S.~Bird, E.~Klein, and E.~Loper, \emph{Natural language processing with Python: analyzing text with the natural language toolkit}.\hskip 1em plus 0.5em minus 0.4em\relax " O'Reilly Media, Inc.", 2009.

\bibitem{cao2023defending}
B.~Cao, Y.~Cao, L.~Lin, and J.~Chen, ``Defending against alignment-breaking attacks via robustly aligned llm,'' \emph{arXiv preprint arXiv:2309.14348}, 2023.

\bibitem{reimers-2019-sentence-bert}
\BIBentryALTinterwordspacing
N.~Reimers and I.~Gurevych, ``Sentence-bert: Sentence embeddings using siamese bert-networks,'' in \emph{Proceedings of the 2019 Conference on Empirical Methods in Natural Language Processing}.\hskip 1em plus 0.5em minus 0.4em\relax Association for Computational Linguistics, 11 2019. [Online]. Available: \url{https://arxiv.org/abs/1908.10084}
\BIBentrySTDinterwordspacing

\bibitem{qi2023fine}
X.~Qi, Y.~Zeng, T.~Xie, P.-Y. Chen, R.~Jia, P.~Mittal, and P.~Henderson, ``Fine-tuning aligned language models compromises safety, even when users do not intend to!'' \emph{arXiv preprint arXiv:2310.03693}, 2023.

\bibitem{huang2023catastrophic}
Y.~Huang, S.~Gupta, M.~Xia, K.~Li, and D.~Chen, ``Catastrophic jailbreak of open-source llms via exploiting generation,'' \emph{arXiv preprint arXiv:2310.06987}, 2023.

\bibitem{hfdatasets}
``Hugging face datasets,'' \url{https://huggingface.co/datasets}, 2024, accessed: 2024-11-09.

\bibitem{ding2023wolf}
P.~Ding, J.~Kuang, D.~Ma, X.~Cao, Y.~Xian, J.~Chen, and S.~Huang, ``A wolf in sheep's clothing: Generalized nested jailbreak prompts can fool large language models easily,'' \emph{arXiv preprint arXiv:2311.08268}, 2023.

\bibitem{chao2023jailbreaking}
P.~Chao, A.~Robey, E.~Dobriban, H.~Hassani, G.~J. Pappas, and E.~Wong, ``Jailbreaking black box large language models in twenty queries,'' 2023.

\bibitem{xiong2024defensive}
C.~Xiong, X.~Qi, P.-Y. Chen, and T.-Y. Ho, ``Defensive prompt patch: A robust and interpretable defense of llms against jailbreak attacks,'' \emph{arXiv preprint arXiv:2405.20099}, 2024.

\bibitem{deng2024masterkey}
G.~Deng, Y.~Liu, Y.~Li, K.~Wang, Y.~Zhang, Z.~Li, H.~Wang, T.~Zhang, and Y.~Liu, ``Masterkey: Automated jailbreaking of large language model chatbots,'' in \emph{Proc. ISOC NDSS}, 2024.

\bibitem{chang2024play}
Z.~Chang, M.~Li, Y.~Liu, J.~Wang, Q.~Wang, and Y.~Liu, ``Play guessing game with llm: Indirect jailbreak attack with implicit clues,'' \emph{arXiv preprint arXiv:2402.09091}, 2024.

\bibitem{zeng2024autodefense}
Y.~Zeng, Y.~Wu, X.~Zhang, H.~Wang, and Q.~Wu, ``Autodefense: Multi-agent llm defense against jailbreak attacks,'' \emph{arXiv preprint arXiv:2403.04783}, 2024.

\bibitem{metallamaguard2}
L.~Team, ``Meta llama guard 2,'' \url{https://github.com/meta-llama/PurpleLlama/blob/main/Llama-Guard2/MODEL_CARD.md}, 2024.

\bibitem{dubey2024llama3herdmodels}
\BIBentryALTinterwordspacing
A.~.~M. Llama~Team, ``The llama 3 herd of models,'' 2024. [Online]. Available: \url{https://arxiv.org/abs/2407.21783}
\BIBentrySTDinterwordspacing

\bibitem{barrett2023identifying}
C.~Barrett, B.~Boyd, E.~Bursztein, N.~Carlini, B.~Chen, J.~Choi, A.~R. Chowdhury, M.~Christodorescu, A.~Datta, S.~Feizi \emph{et~al.}, ``Identifying and mitigating the security risks of generative ai,'' \emph{Foundations and Trends{\textregistered} in Privacy and Security}, vol.~6, no.~1, pp. 1--52, 2023.

\bibitem{fu-etal-2024-safety}
\BIBentryALTinterwordspacing
Y.~Fu, Y.~Li, W.~Xiao, C.~Liu, and Y.~Dong, ``Safety alignment in {NLP} tasks: Weakly aligned summarization as an in-context attack,'' in \emph{Proceedings of the 62nd Annual Meeting of the Association for Computational Linguistics (Volume 1: Long Papers)}, L.-W. Ku, A.~Martins, and V.~Srikumar, Eds.\hskip 1em plus 0.5em minus 0.4em\relax Bangkok, Thailand: Association for Computational Linguistics, Aug. 2024, pp. 8483--8502. [Online]. Available: \url{https://aclanthology.org/2024.acl-long.461}
\BIBentrySTDinterwordspacing

\bibitem{ji2024beavertails}
J.~Ji, M.~Liu, J.~Dai, X.~Pan, C.~Zhang, C.~Bian, B.~Chen, R.~Sun, Y.~Wang, and Y.~Yang, ``Beavertails: Towards improved safety alignment of llm via a human-preference dataset,'' \emph{Advances in Neural Information Processing Systems}, vol.~36, 2024.

\bibitem{dubey2024llama}
A.~Dubey, A.~Jauhri, A.~Pandey, A.~Kadian, A.~Al-Dahle, A.~Letman, A.~Mathur, A.~Schelten, A.~Yang, A.~Fan \emph{et~al.}, ``The llama 3 herd of models,'' \emph{arXiv preprint arXiv:2407.21783}, 2024.

\bibitem{llama3modelcard}
\BIBentryALTinterwordspacing
AI@Meta, ``Llama 3 model card,'' 2024. [Online]. Available: \url{https://github.com/meta-llama/llama3/blob/main/MODEL_CARD.md}
\BIBentrySTDinterwordspacing

\bibitem{moderation}
``Openai moderation api,'' \url{https://platform.openai.com/docs/guides/moderation}, 2024, accessed: 2024-11-10.

\bibitem{jain2023baseline}
N.~Jain, A.~Schwarzschild, Y.~Wen, G.~Somepalli, J.~Kirchenbauer, P.-y. Chiang, M.~Goldblum, A.~Saha, J.~Geiping, and T.~Goldstein, ``Baseline defenses for adversarial attacks against aligned language models,'' \emph{arXiv preprint arXiv:2309.00614}, 2023.

\bibitem{zheng2024prompt}
C.~Zheng, F.~Yin, H.~Zhou, F.~Meng, J.~Zhou, K.-W. Chang, M.~Huang, and N.~Peng, ``On prompt-driven safeguarding for large language models,'' in \emph{Forty-first International Conference on Machine Learning}, 2024.

\bibitem{pisano2023bergeron}
M.~Pisano, P.~Ly, A.~Sanders, B.~Yao, D.~Wang, T.~Strzalkowski, and M.~Si, ``Bergeron: Combating adversarial attacks through a conscience-based alignment framework,'' \emph{arXiv preprint arXiv:2312.00029}, 2023.

\bibitem{zeng2024shieldgemma}
W.~Zeng, Y.~Liu, R.~Mullins, L.~Peran, J.~Fernandez, H.~Harkous, K.~Narasimhan, D.~Proud, P.~Kumar, B.~Radharapu \emph{et~al.}, ``Shieldgemma: Generative ai content moderation based on gemma,'' \emph{arXiv preprint arXiv:2407.21772}, 2024.

\bibitem{xu2024comprehensive}
Z.~Xu, Y.~Liu, G.~Deng, Y.~Li, and S.~Picek, ``A comprehensive study of jailbreak attack versus defense for large language models,'' in \emph{Findings of the Association for Computational Linguistics ACL 2024}, 2024, pp. 7432--7449.

\bibitem{ren2024learning}
M.~Ren, B.~Cao, H.~Lin, C.~Liu, X.~Han, K.~Zeng, G.~Wan, X.~Cai, and L.~Sun, ``Learning or self-aligning? rethinking instruction fine-tuning,'' \emph{arXiv preprint arXiv:2402.18243}, 2024.

\bibitem{zhao2024long}
H.~Zhao, M.~Andriushchenko, F.~Croce, and N.~Flammarion, ``Long is more for alignment: A simple but tough-to-beat baseline for instruction fine-tuning,'' \emph{arXiv preprint arXiv:2402.04833}, 2024.

\bibitem{bai2022training}
Y.~Bai, A.~Jones, K.~Ndousse, A.~Askell, A.~Chen, N.~DasSarma, D.~Drain, S.~Fort, D.~Ganguli, T.~Henighan \emph{et~al.}, ``Training a helpful and harmless assistant with reinforcement learning from human feedback,'' \emph{arXiv preprint arXiv:2204.05862}, 2022.

\bibitem{ethayarajh2024kto}
K.~Ethayarajh, W.~Xu, N.~Muennighoff, D.~Jurafsky, and D.~Kiela, ``Kto: Model alignment as prospect theoretic optimization,'' \emph{arXiv preprint arXiv:2402.01306}, 2024.

\bibitem{zhu2023autodan}
S.~Zhu, R.~Zhang, B.~An, G.~Wu, J.~Barrow, Z.~Wang, F.~Huang, A.~Nenkova, and T.~Sun, ``Autodan: Automatic and interpretable adversarial attacks on large language models,'' \emph{arXiv preprint arXiv:2310.15140}, 2023.

\bibitem{yu2024llm}
J.~Yu, X.~Lin, Z.~Yu, and X.~Xing, ``$\{$LLM-Fuzzer$\}$: Scaling assessment of large language model jailbreaks,'' in \emph{33rd USENIX Security Symposium (USENIX Security 24)}, 2024, pp. 4657--4674.

\bibitem{ge2023mart}
S.~Ge, C.~Zhou, R.~Hou, M.~Khabsa, Y.-C. Wang, Q.~Wang, J.~Han, and Y.~Mao, ``Mart: Improving llm safety with multi-round automatic red-teaming,'' \emph{arXiv preprint arXiv:2311.07689}, 2023.

\bibitem{phute2023llm}
M.~Phute, A.~Helbling, M.~Hull, S.~Peng, S.~Szyller, C.~Cornelius, and D.~H. Chau, ``Llm self defense: By self examination, llms know they are being tricked,'' \emph{arXiv preprint arXiv:2308.07308}, 2023.

\bibitem{li2023rain}
Y.~Li, F.~Wei, J.~Zhao, C.~Zhang, and H.~Zhang, ``Rain: Your language models can align themselves without finetuning,'' \emph{arXiv preprint arXiv:2309.07124}, 2023.

\end{thebibliography}
